\pdfoutput=1
\documentclass[acmtog]{acmart}

\AtBeginDocument{%
  }

\setcopyright{acmlicensed}
\copyrightyear{2026}
\acmYear{2026}
\acmDOI{XXXXXXX.XXXXXXX}

\acmJournal{TOG}

\usepackage{multirow}
\usepackage[ruled]{algorithm2e} 
\usepackage[table]{xcolor}
\usepackage{wrapfig}

\usepackage{color}
\usepackage{graphicx}
\usepackage{booktabs}
\usepackage{amsmath}
\usepackage{array}
\usepackage{enumitem}
\usepackage{gensymb}
\usepackage{float}

\definecolor{green}{rgb}{0, 0.5, 0}
\definecolor{orange}{rgb}{0.8, 0.6, 0.2}
\definecolor{red}{rgb}{1.0, 0.0, 0.0}
\definecolor{teal}{rgb}{0.0, 0.4, 0.4}
\definecolor{purple}{rgb}{0.65,0,0.65}
\definecolor{saffron}{rgb}{0.95,0.75,0.2}
\definecolor{turquoise}{rgb}{0.0,0.5,0.5}
\definecolor{black}{rgb}{0.0, 0.0, 0.0}
\definecolor{gray}{rgb}{0.5, 0.5, 0.5}

\newcommand{\rzz}[1]{{\color{black}#1}}
\newcommand{\sn}[1]{{\color{black}#1}}
\newcommand{\nt}[1]{{\color{black}#1}}

\newcommand{\rev}[1]{{\color{black}#1}}

\begin{document}

\title{Articulate That Object Part (ATOP): 3D Part Articulation from Text and via Motion Personalization}

\author{Aditya Vora}

\affiliation{%
  \institution{Computing Science, Simon Fraser University}
  \city{Burnaby}
  \state{British Columbia}
  \country{Canada}
}
\email{ava40@sfu.ca}

\author{Sauradip Nag}
\affiliation{%
  \institution{Computing Science, Simon Fraser University}
  \city{Burnaby}
  \country{Canada}}
\email{snag@sfu.ca}

\author{Kai Wang}
\affiliation{%
  \institution{Computing Science, Simon Fraser University}
  \city{Burnaby}
  \state{British Columbia}
  \country{Canada}
}
\email{kwang.ether@gmail.com}

\author{Hao (Richard) Zhang}
\affiliation{%
  \institution{Computing Science, Simon Fraser University}
  \city{Burnaby}
  \country{Canada}}
\email{haoz@sfu.ca}

\begin{abstract}
We present ATOP (Articulate That Object Part), a novel {\em few-shot\/} method based on {\em motion personalization\/} to articulate a static 3D object with respect to a part and its motion as prescribed in a text prompt. Given the scarcity of available datasets with motion attribute annotations, existing methods struggle to generalize well in this task. In our work, the text input allows us to tap into the power of modern-day diffusion models to generate plausible motion samples for the right object category and part. In turn, the input 3D object provides ``{\em image prompting\/}'' to personalize the generated motion to the very input object. Our method starts with a few-shot finetuning \nt{to inject \rzz{articulation} awareness to current diffusion models to learn a unique motion identifier associated with the \rzz{target object part}.} \rzz{Our finetuning is applied to a} pre-trained diffusion model for controllable multi-view motion generation, \rzz{trained with} a small collection of \nt{reference motion frames demonstrating \rzz{appropriate} part motion. \rzz{The resulting} motion model can then be \rzz{employed} to realize plausible motion of the input 3D object from multiple views.} At last, we transfer the personalized motion to the 3D space of the object via differentiable rendering to optimize part \rzz{articulation} parameters by a score distillation sampling loss. Experiments on PartNet-Mobility and ACD datasets demonstrate that our method \rzz{can generate realistic motion samples with higher accuracy, leading to} more generalizable 3D motion predictions compared to prior approaches in the few-shot setting.

\end{abstract}




\begin{teaserfigure}
    \centering
    \includegraphics[width=\textwidth]{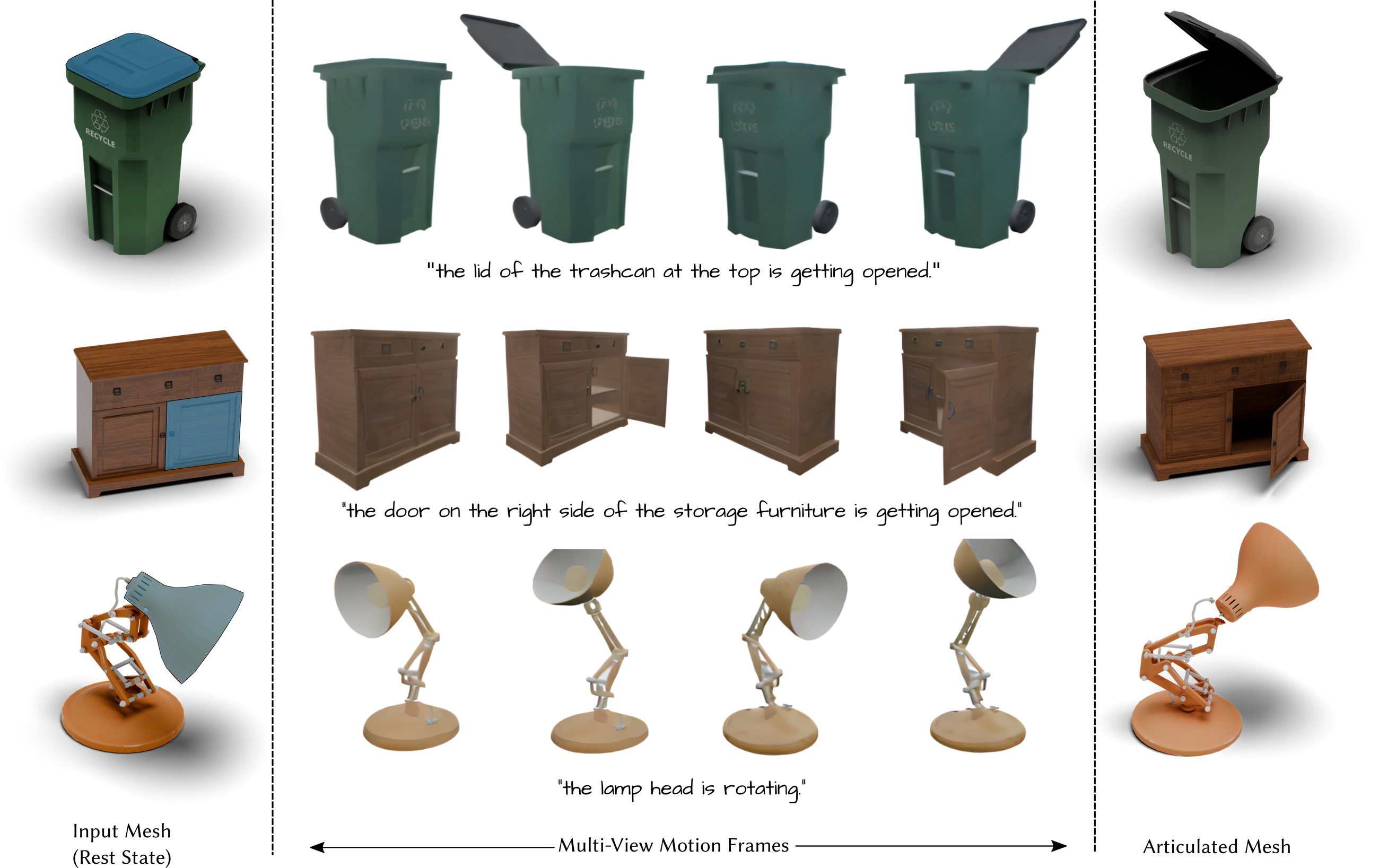}
    \caption{Given a textured mesh in rest state (left), the part to articulate (highlighted), and a text prompt \rzz{describing part motion} as input, we generate multi-view motion frames (middle) of plausible part articulations \rzz{specific to the input object} using a personalized motion diffusion model. We then transfer the motion from the generated frames to the 3D mesh (right) to obtain articulation parameters. Note that the input meshes are the {\em top retrieval results\/} from Objaverse \cite{deitke2024objaverse} for the ``Trash can", "Cabinet" and ``Lamp" \rzz{categories, i.e., not cherry-picked,} showcasing the generalization ability of our approach.}
    \Description{Teaser figure showing a mesh in rest state, highlighted part, text prompt, generated multi-view motion frames, and resulting articulated mesh.}
    \label{fig:teaser}
\end{teaserfigure}

\maketitle

\section{Introduction}
\label{sec:intro}

Everyday objects in the 3D world we live in undergo various movements and articulations. The ability to model and reason about object articulations in 3D plays an important role in simulation, design, autonomous systems, as well as robotic vision and manipulation.
Over the past ten years, there has been a steady build-up of digital 3D assets, from ShapeNet~\cite{chang2015shapenet} (3M models)  in the early days, to efforts on improving the assets' structure (e.g., PartNet~\cite{mo2019partnet} with 27K finely segmented models) and quality (e.g., Amazon Berkeley Objects (ABO)~\cite{collins2022abo} with $\sim$8K models), and to the latest and largest open-source repository Objaverse-XL~\cite{deitke2023objaversexluniverse10m3d} (10M+ models).

However, very few, if any, of these models come with part motions. Predominantly, they were all constructed in their rest (i.e., unarticulated) states. To our knowledge, the largest 3D datasets with part articulations, PartNet-Mobility~\cite{xiang2020sapien} and 
Shape2Motion~\cite{wang2019shape2motion}, only contain 2,346 and 2,440 {\em manually\/} annotated synthetic models, respectively. \rzz{In the more recent dataset Objaverse ~\cite{deitke2023objaverse}, a subset of about
$40k$ models are labeled as ``dynamic,'' as each consists of a series of meshes representing frames of an animation. However, these models do not possess motion attribute annotations.} Clearly, human annotations are expensive in terms of time, cost, and expertise and are not scalable to endow large volumes of 3D assets with articulations. 

In this paper, we wish to develop a learning method to generate {\em plausible\/} and {\em accurate\/} articulations for {\em any\/} given 3D object, {\em without\/} relying on any human-annotated 3D part motions as most existing methods~\cite{hu2017learning,wang2019shape2motion,abbatematteo2019learning,mo2021where2act,jain2021screwnet,yan2020rpm,jiang2022opd} do. In addition, we would also like to solve this problem in a {\em few-shot\/} setting, making the task even more challenging. 


With the recent success of {\em video diffusion models\/}~\cite{blattmann2023stable,wang2024vc,xing2024dynamicrafter,guo2023animatediff,wan2025wan,yang2024cogvideox,li2025puppet}, a natural next step is to use their zero-shot and open vocabulary capabilities to generate motion samples for a target object specifically using an {\em Image-to-Video} (I2V) diffusion model. These motion samples can then be passed to a generalized {\em Video-to-Multi-View video} generator (V2MV) like SV4D \cite{xie2024sv4d, yao2025sv4d} to produce motion from multiple viewpoints. The resulting multi-view motion can be used to perform motion transfer from multiple views to 3D. This method does not require 3D annotations, making it more scalable, but it comes with \rzz{three} key challenges for achieving accurate 3D part articulations:
\begin{figure*}[!t]
   \includegraphics[width=\textwidth]{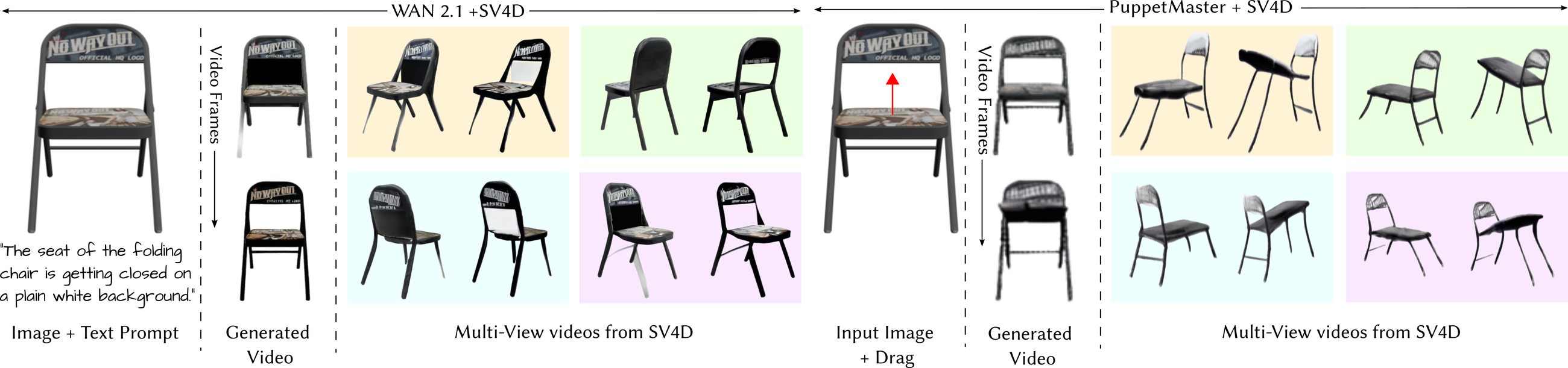}
    \caption{
        We conduct two experiments using recent I2V models to generate articulation motion. First, we use one of the SOTA I2V diffusion model WAN 2.1 \cite{wan2025wan} with a text prompt and image (Fig. left) to generate a video, followed by SV4D \cite{xie2024sv4d, yao2025sv4d} for multi-view motion generation. Due to limited control via text, WAN 2.1 fails to produce accurate articulation, which leads to errors in SV4D outputs. Second, we test another I2V diffusion model named PuppetMaster \cite{li2025puppet} with a “drag” prompt (Fig. right). However, because of drag’s ambiguity in determining parts, it causes the model to lift the chair seat instead of folding it, again resulting in implausible SV4D reconstructions.
        }
    \Description{Comparison of articulation motion generated by WAN 2.1 and PuppetMaster models, showing differences in accuracy and plausibility of the resulting motions.}

    \vspace{-3mm}
\label{fig:diff_articulation}
\end{figure*}


\begin{itemize}[leftmargin=*]

\item First, while SOTA video diffusion models 
can achieve unprecedented visual quality for open-domain text-to-video \rzz{and I2V} generations, they are not yet well trained or sufficiently adaptive to generate articulations of common everyday objects, especially when the typical {\em piecewise rigidity\/} is expected. \rzz{Fig. \ref{fig:diff_articulation} (left) shows such an example, with} videos generated \rzz{by a state-of-the-art I2V diffusion model WAN2.1 \cite{wan2025wan}, from an input image of a folding chair using a motion-descriptive} text prompt.

\item
Second, \rzz{as Fig.~\ref{fig:diff_articulation} (left) further shows, even with SOTA diffusion models, input from text prompts only is too coarse and often fails to control the precise movements of individual object parts in an image, which is crucial for articulation estimation.}

More promising are recent \rzz{motion models under explicit control, such as ``drag" interactions}~\cite{li2025puppet}. However, these models still do not consistently produce plausible video outputs \rzz{due to two reasons}. First, \rzz{the cost of training is significantly higher than for I2V diffusion with text inputs, e.g., due to the need for manually annotated drag-based ground truths.} 
Moreover, these approaches often suffer from ambiguity in interpreting the drag signals, especially over multiple object parts, making it difficult to accurately identify and animate an intended part, as seen in Fig. ~\ref{fig:diff_articulation} (right) where the chair seat was lifted rather than folded.

\item
Third, \rzz{lifting motions from a single view to multiple views is challenging, but essential for faithfully transferring the articulation from 2D to 3D space.} As shown in Fig. ~\ref{fig:diff_articulation} (right), SV4D struggles to produce plausible multi-view motion samples directly. This often leads to significant structural distortions, for instance, altering the shape of the chair which ultimately introduces errors in the final articulation estimation in 3D space.
\end{itemize}

\begin{figure*}
\vspace{-1em}
    \includegraphics[width=\textwidth]{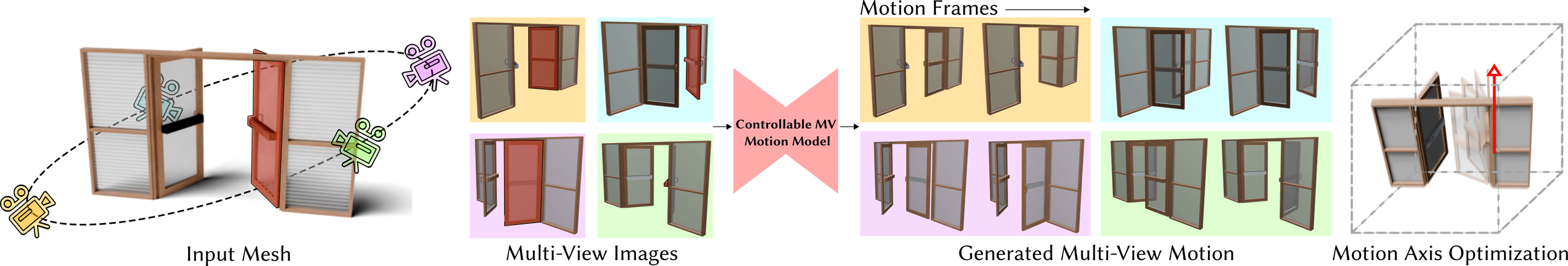}
\caption{{\em Inference Pipeline\/}: As the first step of our inference pipeline, we take a textured segmented mesh as input. If part segmentation is not available, we apply off-the-shelf open-vocabulary methods such as Part-SLiP \cite{liu2023partslip} or PartSTaD \cite{kim2024partstad} to obtain part segments. We then render multi-view images and masks of this mesh and pass it through a controllable multi-view motion model which hallucinates part motion for given camera poses which is personalized to the target input shape. Different background color of images indicates different views. Once this motion samples are obtained we then transfer the motion to 3D space by directly optimizing for the motion-axis and origin.}
\label{fig:pipeline}
\end{figure*}

\rzz{To address these challenges}, we introduce a novel \rzz{{\em few-shot\/}} method for articulating static 3D objects based on {\em motion personalization\/} guided by a controllable, part-aware, and multi-view motion generator. \rzz{Specifically, the input to our method consists of a mesh model with a masked part that is to be articulated, whose motion is prescribed in a text prompt. We aim for a lightweight}
model design for \nt{multi-view motion generation} that, once trained on a few reference samples of \rzz{the desirable} part motion, can directly hallucinate motion patterns \rzz{as controlled by the masked part from multiple views, and personalized to the input object.} This is \rzz{facilitated by the} text prompt and multi-view images rendered from the input mesh; see Figs.~\ref{fig:teaser} and~\ref{fig:pipeline}. 
Our method is coined ATOP, for Articulate That Object Part, and it consists of two main \rzz{steps/stages}:


\begin{enumerate}[itemsep=2pt, parsep=0pt, topsep=2pt, leftmargin=*]
\item {\em \rzz{Few-shot} finetuning for \rzz{personalized, multi-view} part motion generation.\/} 
This is the key first step to compensate for the lack of articulation awareness by current diffusion models. 

We finetune a pre-trained multi-view \textit{image} generator, ImageDream~\cite{wang2023imagedream}, for \rzz{mask-controlled} multi-view motion generation, using few-shot motion samples available during training. Specifically, we start from the pre-trained weights of ImageDream and inflate them for video processing, with motion region-of-interest and multi-view controls enabled through text and camera pose conditioning. Once finetuned, given a new target 3D mesh, we render multi-view images and masks of the part to be articulated, and these images and masks are fed to the diffusion model which hallucinates the desired part motion from multiple views, personalized to the target object; see Fig.~\ref{fig:pipeline}.

\item {\em Multiview-to-mesh motion transfer.\/} Using the personalized motion output from the previous stage, we transfer the plausible motion to the target part on the input 3D object via differentiable rendering, which optimizes the motion axis parameters associated with the prescribed part by a score distillation sample (SDS) loss \cite{poole2022dreamfusion}.
\end{enumerate}

\rzz{Compared to using drag-controlled motion generation~\cite{li2025puppet}, part masks are more precise in disambiguating motion targets and they are easy to obtain using SOTA segmentation models, e.g. SAM ~\cite{ravi2024sam2segmentimages}, avoiding the need for human intervention.} 
%
\rzz{Also worth noting is that our motion generation in the first stage is performed in a {\em single\/} pass to directly produce multi-view motions from images. We demonstrate experimentally that this is less prone to error propagation than a straightforward two-stage pipeline involving first I2V and then V2MV generation.}


To our knowledge, ATOP represents the first annotation-free learning framework for generating \rzz{plausible and} accurate part articulations on static 3D objects from text prompts. We present qualitative and quantitative experiments to show that our method is not only capable of generating realistic and personalized multi-view motions on unseen shapes, but can also effectively transfer the motions onto 3D meshes, resulting in more accurate \rzz{articulation prediction and realization} than prior works in the few-shot setting.

\section{Related Work}

\label{sec:related work}

\noindent\textbf{Mobility analysis.} Understanding part mobility is essential for kinematics \cite{abbatematteo2019learning, mo2021where2act}. Recent data-driven approaches leverage 3D datasets with articulation annotations to predict motion attributes. ScrewNet \cite{jain2021screwnet} infers articulations directly from depth image sequences without requiring part segmentation, while Hu et al. \cite{hu2017learning} classify motion types from point clouds. Similarly, Shape2Motion \cite{wang2019shape2motion}, RPM-Net \cite{yan2020rpm}, and OPD \cite{jiang2022opd} jointly predict motion-oriented part segmentation and attributes from point clouds and images. While these approaches enable fast inference, their reliance on extensive human annotations limits scalability and generalizibility. In contrast, our method piggybacks on diffusion foundational models to generalize under a few-shot setting. 

Another line of works estimates motion attributes of an object as a by-product in the process of reconstructing digital twins of a scanned object \cite{jiang2022ditto, liu2023paris, wei2022nasam, song2024reacto, mandi2024real2code}. However, different from our work, these methods focus on high-fidelity reconstruction of the scanned object from multi-view images, monocular videos or depth maps where the scan is captured in {\em different articulation} states assuming the object is already articulated. In contrast to these approaches, our work focuses on estimating plausible motion axis parameters for a {\em static mesh}, hence enabling dynamic attributes.

\vspace{3pt}

\noindent\textbf{Motion personalization.} Diffusion models~\citep{ho2020denoising, song2020denoising, song2020score} have garnered significant interest for their training stability and remarkable performance in text-to-image (T2I) generation~\citep{ramesh2021zero, balaji2022ediffi}. Video generation~\citep{le2021ccvs, yu2023magvit, luo2023videofusion} extends image generation to sequential frames, incorporating temporal dynamics. Recent advances~\citep{singer2022make, zhou2022magicvideo, ge2023preserve, nag2023difftad, alimohammadi2024smite} adapt T2I diffusion to spatio-temporal domains via architectural modifications. Personalizing motions using reference videos has gained traction, with methods such as Tune-a-Video~\citep{tuneavideo} enabling one-shot video editing with structural control. Current approaches~\citep{materzynska2023customizing, jeong2023vmc} refine pre-trained text-to-video (T2V) models using regularization or frame residual losses, reducing dependency on training video appearances. Subject-driven video generation~\citep{dreamix, wu2023lamp, zhao2024motiondirector} fine-tunes video diffusion models but often struggles with overfitting and limited dynamics, failing to accurately capture 3D articulations.
\sn{Recent advances in video diffusion models like WAN\cite{wan2025wan} and CogVideo \cite{hong2022cogvideo} can generate global object motions but these methods do not allow controlling motion at the level of object parts, making it hard to generate part motions at desired spatial locations .}
In contrast, our method proposes a motion personalization approach that enables controllable multi-view video generation, enabling part motions at desired spatial locations, which is crucial for articulation.

\noindent\textbf{Generative articulation.} With the rapid advancements in generative AI in recent years, the task of learning to represent, reconstruct, and generate object articulations from various input sources has become a prominent research area in visual computing~\cite{li2025puppet,uzolas2024motiondreamer}. NAP~\cite{lei2023nap} employed diffusion models to synthesize articulated 3D objects by introducing a novel articulation graph parameterization. However, it faced limitations in scalability and controllability for articulations. CAGE~\cite{liu2024cage} addressed these challenges by jointly modeling object parts and their motions within a graph-based structure. Recently, \cite{liu2024singapo} proposed the seminal task of generating articulated objects from a single image. In contrast to these existing approaches that are focused on generating articulated assets, our goal is to enhance existing 3D assets by incorporating precise articulations.

Concurrent to our work, Articulate Any Mesh (AAM) \cite{qiu2025articulate} constructs a ``digital twin" of an input 3D mesh with articulation, instead of directly inferring motion parameters on the input mesh as our work. For articulation generation, AAM leverages GPTo’s foundational knowledge on {\em joint classification} only, over the types of prismatic or revolute joints, while the joint locations and articulation axes are both estimated through heuristic shape analysis. 
With ATOP, we optimize motion axis directions by aligning them with plausible video outputs generated during the motion personalization step. This ensures that the target part is articulated correctly with proper motion axis orientations.

\vspace{3pt}

\noindent\textbf{Video/Multi-View Video Generation.}
\sn{Building upon the success of large-scale text-to-image diffusion models \cite{saharia2022photorealistic, podell2023sdxl}, recent advancements have extended diffusion frameworks to encompass video and 3D generation. Video synthesis models such as AnimateDiff \cite{guo2023animatediff}, DynamiCrafter \cite{xing2024dynamicrafter}, and recent models like \cite{wan2025wan, yang2024cogvideox}, incorporate temporal dynamics while preserving the visual fidelity of their text-to-image diffusion models albeit often without controllable camera motion. To overcome this limitation, camera-conditioned multi-view video generation methods \cite{zeng2024stag4d, nag20252, yao2025sv4d, ren2024l4gm, zhang20244diffusion, li2024vivid} leverage 3D-aware attention mechanisms from multi-view image generation models \cite{wang2023imagedream, voleti2025sv3d} to ensure view-consistency, while employing self-attention expansion for temporal coherence. These methods typically rely on explicit dynamic supervision from Objaverse-Animation sequences \cite{deitke2023objaverse}. However, they generally require monocular video or textual input and lack fine-grained part-aware motion control, rendering them unsuitable for animating a given 3D mesh. Concurrent to our work, PuppetMaster \cite{li2025puppet} enables controllable video generation with part dynamics from a single image using user-provided ``drag” signals. However, this approach faces limited generalizibility due to its dependence on large-scale human annotations for ``drag" inputs as well as limited dynamic 3D datasets. Moreover as shown in our experiments, drag-based control is not always reliable, particularly in cases of part ambiguity, often resulting in erroneous outputs. In contrast, ATOP presents a unified framework that generates part-aware, controllable multi-view motion videos from {\em multi-view\/} images and corresponding {\em part masks\/} extracted from a mesh. These part masks effectively disambiguate the ROI, enabling more accurate and controllable motion generation.}

\section{Preliminaries}
\label{preliminaries}

\begin{figure*}
\centering
\includegraphics[width=\textwidth]{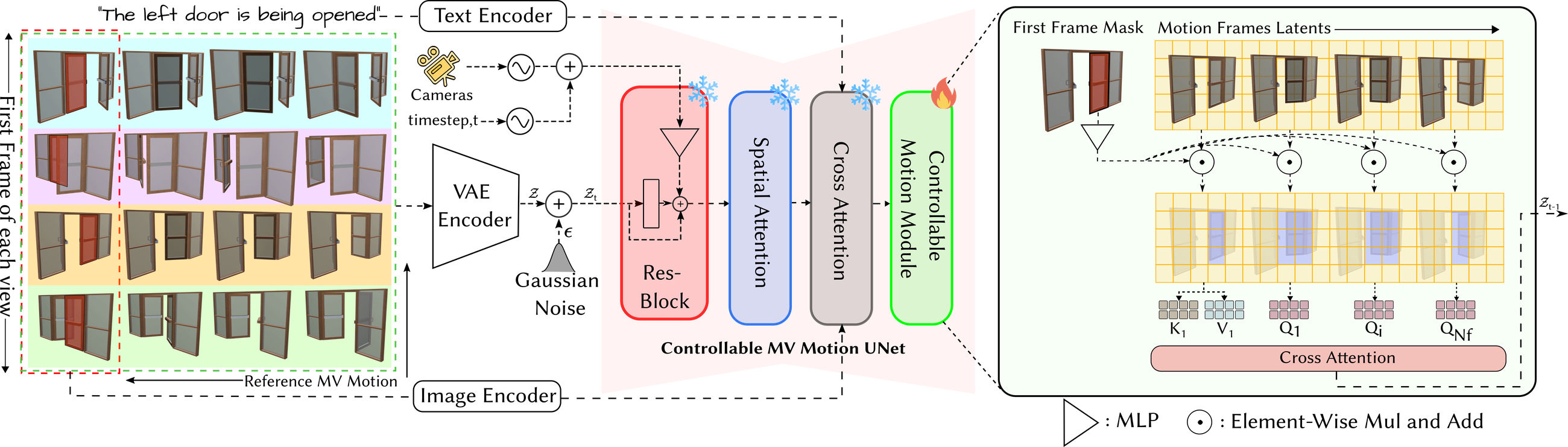}

\caption{{\em Training Pipeline for Multi-View Motion Generation via Personalization:\/} Reference {\em multi-view\/} motion videos (each view shown in a different color) are first encoded using a VAE encoder. Noise is then added to the encoded features, which are subsequently passed through a {\em controllable multi-view\/} motion U-Net. To capture the unique motion pattern in the reference video, we introduce a {\em trainable\/} and {\em controllable\/} motion module that learns both the motion pattern {\em and\/} the spatial controllability of motion from the reference motion samples and the first-frame masks available during training. For simplicity, we illustrate the temporal processing block using only a single view (shown in yellow), whereas in the actual implementation, temporal attention is applied across all views.}

\vspace{-4mm}
\label{fig:pipeline-training}
\end{figure*}

\noindent\textbf{Inflated UNet.} A Text2Image (T2I) diffusion model, such as LDM~\citep{rombach2022high}, typically adopts a U-Net architecture~\citep{ronneberger2015u}, which comprises of a downsampling phase followed by upsampling with skip connections. This architecture is structured using stacked 2D convolutional residual blocks, spatial attention blocks, and cross-attention blocks that integrate embeddings derived from textual prompts. To adapt the T2I model for Text2Video (T2V) tasks, the convolutional residual blocks and spatial attention blocks are modified through inflation. Consistent with prior work~\citep{FLATTEN,tuneavideo}, the $3 \times 3$ convolution kernels in the residual blocks are replaced with $1 \times 3 \times 3$ kernels by introducing a pseudo-temporal channel. To improve temporal coherence, the spatial self-attention mechanism is also extended to the spatio-temporal domain. While the original spatial self-attention mechanism operated on patches within individual frames, the inflated U-Net leverages patch embeddings across the entire video as queries, keys, and values. This design facilitates a comprehensive understanding of the video context. Furthermore, the parameters from the original spatial attention blocks are reused in the newly designed dense spatio-temporal attention blocks.

\noindent\textbf{Score Distillation Sampling.} It is a method that enables using a diffusion model as a critic, i.e., using it as a loss without explicitly back-propagating through the diffusion process. To perform score distillation, noise is first added to a given image (e.g., one novel view). Then, the diffusion model is used to predict the added noise from the noised image. Finally, the difference between the predicted and added noises is used for calculating per-pixel gradients. The per-pixel score distillation gradients is given as: 
\begin{equation}
    \nabla_{\theta} \mathcal{L}_{SDS}(\mathbf{z}, \tau, \epsilon, t) = \mathbb{E}_{t, \epsilon} \left [ \omega(t)(\epsilon_\phi(z_t; t, \tau) - \epsilon) \frac{\partial \mathbf{z_{t}}}{\partial \theta}\right ]
    \label{eq: sds_prelim}
\end{equation}
where $\theta$ represents the parameters of the 3D representation, $\mathbf{z}$ is the rendered latent at the current view, $\tau$ is the text prompt, $t$ is the timestamp in the diffusion process, $\epsilon$ the ground truth noise, $\epsilon_{\phi}$ is the UNet with frozen parameters $\phi$, whereas $\mathbf{z_t}$ is the latent obtained by adding noise in timestep $t$. During training, the gradients flow from the pixel gradients to the 3D representation.

\section{Method}
\label{method}


\subsection{Problem Setting}
\label{sec:problem setting}


Given a segmented and textured {\em static\/} mesh (e.g., a door) and a motion prompt $\tau$ (e.g., ``The left door is being opened''), our objective is to estimate the motion parameters needed to articulate the specified part of the object. If segmentation is not available, we use off-the-shelf open-vocabulary methods based on vision-language models \cite{liu2023partslip, kim2024partstad} to segment the mesh. We approach the problem of articulation estimation in two main stages: {\em a) Multi-View motion generation via Personalization \/} (Sec.~\ref{sec:ft_for_mv_video-gen}) and {\em b) 3D Motion Axis Optimization \/} (Sec.~\ref{sec: 3D motion inference}). In the first stage, we use a motion prompt $\tau$, part mask $\mathcal{B}$ of the first frame, and a set of reference multi-view motion videos $\mathcal{V}$ to learn the motion dynamics of object parts for a novel shape category. This is done by fine-tuning a pre-trained image-to-multi-view diffusion (I2MV) model \cite{wang2023imagedream} by inflating it for controllable multi-view motion generation. Post finetuning, we can then use this model to hallucinate part motion for any target object from the same novel category. In the second stage, we use the synthesized multi-view motion samples from the diffusion model for the target shape, to estimate the 3D motion axis for the mesh $\mathcal{M}$. Training objectives used for all stages is described in Sec. ~\ref{sec: training_objectives}. An overview of our approach is shown in Fig.~\ref{fig:pipeline} and ~\ref{fig:pipeline-training}.


\subsection{Multi-View Motion Generation via  Personalization} 
\label{sec:ft_for_mv_video-gen}
As shown in Fig. ~\ref{fig:diff_articulation}, a straightforward two-stage pipeline that combines state-of-the-art I2V \cite{wan2025wan, li2025puppet} with V2MV diffusion models like SV4D \cite{xie2024sv4d, yao2025sv4d} for this task often struggles in representing articulated motion in a {\em zero-shot\/} manner. To address this, we introduce a {\em single-stage}, light-weight finetuning step that efficiently incorporates multi-view articulation awareness into a diffusion model while preserving data efficiency and generalizibility. This task is challenging because: (1) Existing {\em finetuning\/} methods for video generation \cite{tuneavideo, qi2023fatezero, zhao2024motiondirector} cannot generate 3D-aware, multi-view videos with generalizibility. (2) Recent advances in text to 4D generation \cite{zhang20244diffusion, li2024vivid} lack spatially controllable motion, which is essential for articulation. To overcome these limitations, we propose a finetuning strategy that adapts an I2MV diffusion model for {\em controllable} {\em multi-view} motion generation {\em directly\/} in a single inference process. Since these models operate on image domain, we extend them to incorporate motion information using a network inflation approach \citep{FLATTEN, tuneavideo}, enabling multi-view motion synthesis. This process is divided into three steps which ensure that the generated samples are: 1) {\em multi-view\/} consistent 2) preserves {\em appearance\/} and {\em structure\/} of the target shape 3) motion is {\em spatially\/} controllable.

\subsubsection{Correspondence-Aware Spatial Attention.} 
\begin{figure}
\centering
   \includegraphics[scale=0.09]{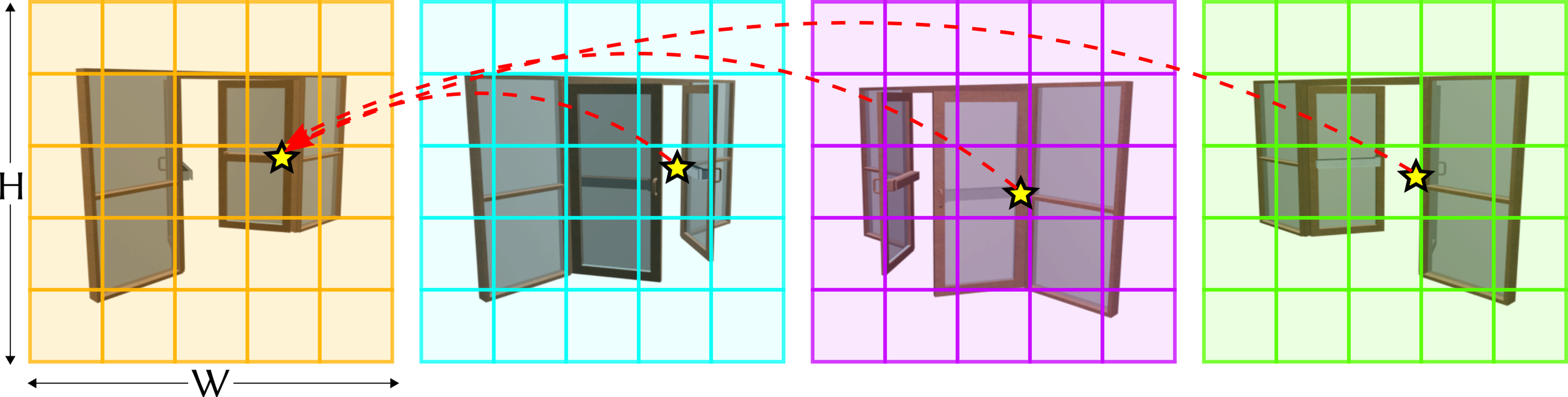}

    \caption{Correspondence-Aware Spatial Attention across multi-view latent. Each color represents different view. All the temporal frames are batched and same operation is applied between corresponding views of each frame.}
\label{fig:corr_self_attention}
\vspace{-3mm}
\end{figure}

We want the generated multi-view motion samples to be geometrically consistent to ensure faithful reconstruction of coarse articulated shape during the 3D motion axis inference step. For this, we inflate the correspondence aware self-attention module of {\em pre-trained\/} Image-Dream model to capture the geometric consistency among the generated multi-view motion samples. Given a multi-view reference motion sample as input during finetuning, we first encode these frames into a $6D$ latent $\text{Z}$ using a VAE encoder $\mathcal{E}$ \cite{vae}, and tokenize the latents to size 
$(\text{B}\times \text{N}_{\text{v}}\times\text{N}_{\text{f}}) \times (\text{H}\times\text{W})\times\text{F}$ before applying self-attention using ImageDream’s pre-trained weights. Fig. \ref{fig:corr_self_attention} demonstrates how information is aggregated across views in the self-attention process. Because self-attention layers are trained to capture geometrical consistency across views, we reshape the tensors accordingly to apply attention across views at each frame index independently. This reshaping operation is given as:
\begin{equation}
\small
    \text{Z}_{\text{out}} = \texttt{reshape}(\text{Z}, (\text{B} \text{N}_{\text{v}} \text{N}_{\text{f}})(\text{H}\text{W})\text{F}\rightarrow(\text{B} \text{N}_{\text{f}})(\text{N}_{\text{v}}\text{H}\text{W})\text{F})
\end{equation}
\noindent Here,  $\text{B}$ is the batch size, $\text{N}_\text{v}$ the number of views, $\text{N}_\text{f}$ the number of frames, $\text{H}$ and $\text{W}$ the spatial dimensions of the latent, and $\text{F}$ the feature dimension. The reshaped tensor  $\text{Z}_{\text{out}}$ is then used in subsequent layers. Post spatial attention block, object appearance is integrated into the model via a cross-attention mechanism.





\subsubsection{Cross Attention for Appearance Injection.} Since our goal is to personalize the motion to the target shape, in addition to ensuring that the generated multi-view motion samples have geometric consistency, we also need to preserve the overall structure of the target shape across multiple views in the generated output. This structural alignment helps produce pixel-accurate outputs that closely match the target mesh, ultimately improving the motion axis optimization during the motion inference step. For this, we leverage the pre-trained CLIP image encoder weights, which are trained along-with ImageDream diffusion model for incorporating images as an additional modality for 3D generation. To make sure that the object appearance is preserved across all the viewpoints, we provide {\em multi-view conditioning\/} through CLIP Image encoder and fuse the information with the latents from the diffusion model. For this, \emph{multiple views (e.g., 4) of the object's rest state} (first frame of reference video shown in Fig. ~\ref{fig:pipeline-training}) are encoded into fixed-size embeddings using a pre-trained CLIP image encoder. These embeddings, along with CLIP text embeddings, are fused into the latent space via cross-attention, ensuring each view's features align with its corresponding latent representation. This is achieved as follows: 
\begin{equation}
\small
{\text{F}}_{\text{out}} = \texttt{Attn}(\text{Q}, \text{K}_{\text{txt}}, \text{V}_{\text{txt}}) + \lambda \cdot \texttt{Attn}(\text{Q}, \text{K}_{\text{img}}, \text{V}_{\text{img}})
\end{equation}
In this setup, $\text{Q}$ represents the query tokens from the latent $\text{Z}$, while $\text{K}_{\text{txt}}$ and $\text{V}_{\text{txt}}$ are the keys and values for the text embeddings. $\text{K}_{\text{img}}$ and $\text{V}_{\text{img}}$ are keys and values of image tokens. $\texttt{Attn(.)}$ denotes the attention mechanism \cite{vaswani2017attention}.

\subsubsection{Controllable Part-Motion Module.} Besides geometric consistency and appearance injection, it is crucial for the diffusion model to learn the temporal part-motion patterns in the provided reference motion videos for fintuning. Improper temporal constraints would break the synchronization among different views and introduce geometric inconsistency. However, simply inserting a temporal attention block within the diffusion model might not serve our purpose because our main goal is to achieve part-level controllability in the motion generation so that the generated motion output can be spatially controlled using the {\em part masks\/} of the target object. To achieve this, we first add a new sparse spatio-temporal attention block that transfers the appearance information from the first frame injected via cross attention to other frames of the video. Next, in order to control the spatial location of the motion, we apply an \emph{adaptive affine transformation} to the model's intermediate features. For this, given a mask input $\mathcal{B}$ and the mapping functions $f(.)$ and $g(.)$ we map the mask to scale $\gamma$ and shift $\beta$ and use them to modulate the latents as $\text{Z}^{'} = \gamma \odot \text{Z} + \beta$. These functions ($f(.)$ and $g(.)$) are implemented using two-layer small MLP blocks. A high-level overview of the controllable motion module is shown in Fig. \ref{fig:pipeline-training} (right). Temporal attention is then applied between the first and the last frame as,
\begin{equation}
\small
    \text{Z} = \texttt{Attn}(\text{Q}_{\text{i}}, \text{K}_{1}, \text{V}_{1})
    \label{eq: temp_attention}
\end{equation}
Here, $\text{Q}_{\text{i}}$ represents the query tokens of the $\text{i}^{th}$ frame, while $\text{K}_{1}$ and $\text{V}_{1}$ are the keys and values from the first frame. The final output, $\text{Z}$, is then fed into the decoder VAE to generate the MV motion output.

\begin{figure}
\includegraphics[scale=0.18]{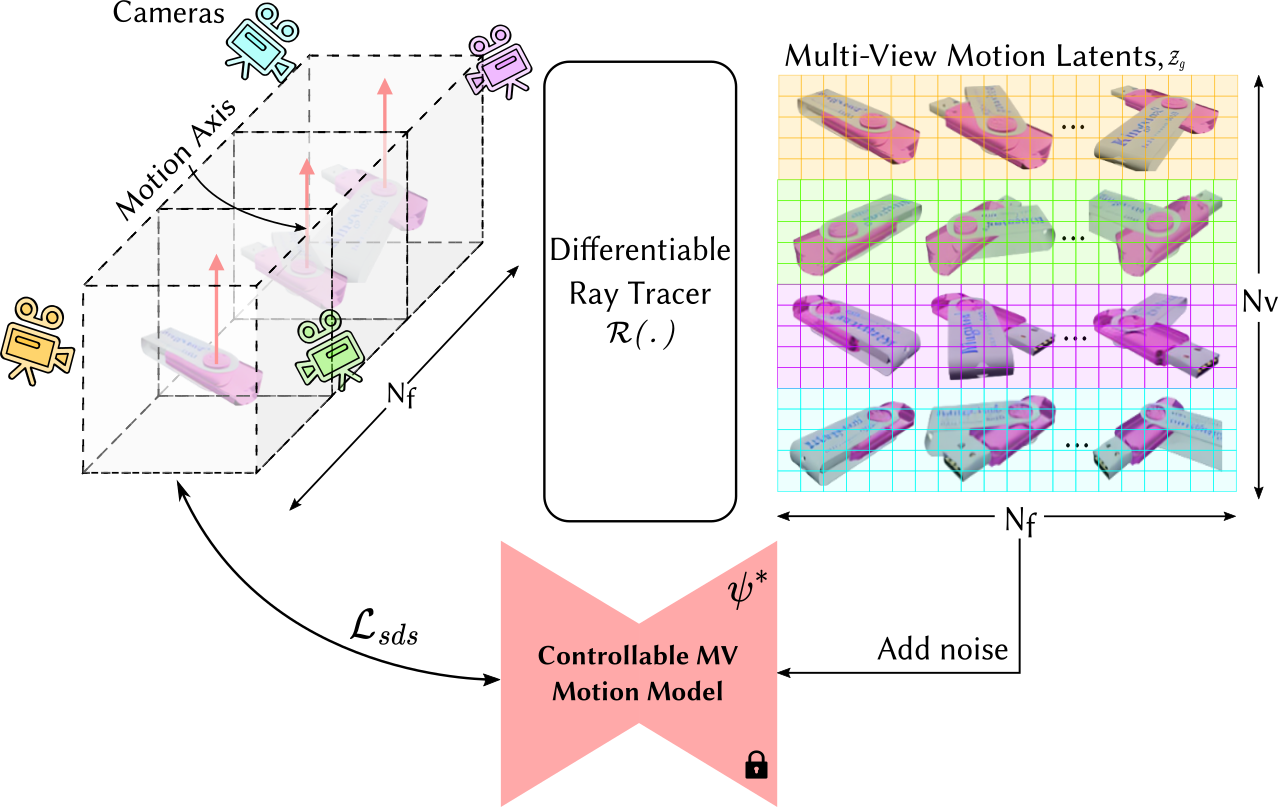}

\caption{Approach of 3D motion axis optimization. We render the multi-view latent $\mathcal{Z}_{g}$ using our differentiable renderer $\mathcal{R}(.)$ and then optimize these latent using the guidance from our personalized motion model $\psi^{*}$.}
\label{fig:pipeline-3d-motion-transfer}
\end{figure}

\subsubsection{Camera Controllability.} For multi-view motion generation, camera poses are incorporated into the diffusion model as input. They are first encoded using MLP layers, transforming them into pose embeddings. These embeddings are then fused with the timestep embeddings and added to the latent after ResBlock as shown in Fig. ~\ref{fig:pipeline-training}.

\subsubsection{Motion Personalization.} We finetune this model using few reference videos available during training and update the newly added part-aware motion module. Post finetuning, we obtain a controllable multi-view motion diffusion model, $\psi^{*}$ that can hallucinate part-motions of any unseen mesh from the same novel category used during training. In order to generate personalized motion to a target mesh, we first render multi-view images $\mathcal{I}_{r}$ and part-masks  $\mathcal{B}$ of the 3D mesh in its rest state using the same fixed camera poses used during finetuning. Then, given a text prompt ($\tau$) and using the rendered multi-view images $\mathcal{I}_{r}$ and part masks $\mathcal{B}$ from camera poses $\mathcal{C}$, we infer the part motion from different views using our personalized diffusion model, $\psi^{*}$ which is given as:
\begin{equation}
\small
    \hat{\mathcal{V}} = \psi^{*}(\mathcal{I}_{r},\tau,\mathcal{B}, \mathcal{C})
    \label{eq: personalization}
\end{equation}
Here, $\mathcal{I}_{r}$ is the rendered image of the target mesh, $\mathcal{B}$ is the corresponding part segmentation mask from camera poses $\mathcal{C}$ and $\tau$ is the text-prompt describing the motion. The generated personalized multi-view motion, $\hat{\mathcal{V}}$, is then used to estimate the 3D motion axis for the target mesh. Refer to Fig. ~\ref{fig:pipeline} for more details.

\begin{table*}
\centering
\begin{tabular}{c|c|ccccccccccc}
\hline
\multirow{2}{*}{\textbf{Metric}}                            & \multirow{2}{*}{\textbf{Method}} & \multicolumn{11}{c}{\textbf{Categories}}                                                                                                                      \\ \cline{3-13} 
                                                   &                         & Dish & Disp & Chair & Micro & Lamp & Oven & Fridge & Storage & Washer & \multicolumn{1}{c|}{Glasses} & Mean \\ \hline
\multirow{3}{*}{\texttt{MAE} $\downarrow$} & S2M \cite{wang2019shape2motion}           &   8.93         &    4.21    &      8.33        &    3.82       &  9.76   &    7.97   &       5.44        &         5.28         &       6.58         & \multicolumn{1}{c|}{ 10.12}      &   7.04  \\
                                                   & OPD \cite{jiang2022opd}                    &     10.65       &     13.83    &      11.52        &    15.37       &  17.27    &   11.45   &    12.27          &        6.32         &        9.34        & \multicolumn{1}{c|}{13.53}      &   12.15   \\
                                                   & Ours-ATOP                &     \cellcolor{blue!15}   \textbf{1.95}      &    \cellcolor{blue!15}\textbf{0.83}     &      \cellcolor{blue!15}\textbf{5.77}        &     \cellcolor{blue!15}\textbf{1.14}      &  \cellcolor{blue!15}\textbf{3.64}   &   \cellcolor{blue!15}\textbf{1.36}   &      \cellcolor{blue!15}\textbf{1.77}        &      \cellcolor{blue!15}\textbf{1.65}            &   \cellcolor{blue!15}\textbf{3.81}             & \multicolumn{1}{c|}{\cellcolor{blue!15}\textbf{4.47}}      &  \cellcolor{blue!15}\textbf{2.63}    \\ \hline
\multirow{3}{*}{\texttt{MPE} $\downarrow$} & S2M \cite{wang2019shape2motion}            &    \cellcolor{blue!15} \textbf{0.12}       &    \cellcolor{blue!15} \textbf{0.11}     &    0.29          &      0.07     &    0.22  &   0.05   &        0.04      &          0.14         &  0.12              & \multicolumn{1}{c|}{ 0.08}      &  0.12    \\
                                                   & OPD \cite{jiang2022opd}                    &    0.21        & 0.17         &      0.36        &    0.09       &    0.18   &   0.08   &     0.08         &        \cellcolor{blue!15}\textbf{0.12}          &      0.11          & \multicolumn{1}{c|}{0.16}      &    0.15  \\
                                                   & Ours - ATOP                  &       0.18      &    0.16     &     \cellcolor{blue!15}  \textbf{0.28}        &   \cellcolor{blue!15}  \textbf{0.03}      &  \cellcolor{blue!15} \textbf{0.13}   &   \cellcolor{blue!15}\textbf{0.03}   &        \cellcolor{blue!15}\textbf{0.03}      &        0.17         &    \cellcolor{blue!15}\textbf{0.04}            &  \multicolumn{1}{c|}{\cellcolor{blue!15}\textbf{0.05}}      &    \cellcolor{blue!15}\textbf{0.11}  \\ \hline
\end{tabular}
\vspace{2mm}
\caption{Quantitative results of Motion Parameter Estimation on PartNet-Mobility Dataset \cite{xiang2020sapien}. We optimize the motion parameters using the multi-view motion output from the diffusion model. All these results show the accuracy of the motion parameters in the 3D space. The results are computed on the test split of each category from \cite{liu2023partslip, kim2024partstad}.}
\label{tab: partnet_sapian_quant}
\end{table*}

\begin{table*}[t!]
\centering
\begin{tabular}{c|c|cccccccc|c}
\hline
\multirow{2}{*}{\textbf{Metric}}                            & \multirow{2}{*}{\textbf{Method}} & \multicolumn{9}{c}{\textbf{Categories}}                                                     \\ \cline{3-11} 
                                                   &                         & Armoire & Cabinet & NightStand & Table & Microwave & Oven & Refrigerator & Washer & Mean \\ \hline
\multirow{3}{*}{\texttt{MAE} $\downarrow$}& S2M \cite{wang2019shape2motion}                    &      17.32   &    19.34     &       22.37     &    21.62   &    23.56       &  22.56    &    25.42    &   19.77     &    21.49  \\
                                                   & OPD \cite{jiang2022opd}       &     12.37    &    9.26     &     10.62       &   8.43    &     16.46      &   14.93   &   13.62     &   11.21     &   12.11   \\
                                                   & Ours - ATOP  &  \cellcolor{blue!15} \textbf{2.83}      &    \cellcolor{blue!15}\textbf{3.29}     &     \cellcolor{blue!15}\textbf{3.92}       &   \cellcolor{blue!15}\textbf{2.39}    &     \cellcolor{blue!15}\textbf{3.52}      &    \cellcolor{blue!15}\textbf{3.34}  &  \cellcolor{blue!15}\textbf{2.96}      &    \cellcolor{blue!15}\textbf{6.75}    &  \cellcolor{blue!15}\textbf{3.63}    \\ \hline
\multirow{3}{*}{\texttt{MPE} $\downarrow$} & S2M \cite{wang2019shape2motion}              &    0.15     &    0.22     &      \cellcolor{blue!15}  \textbf{0.19}    & \cellcolor{blue!15} \textbf{0.16}     &      0.16     &   0.13   &   0.15    &    0.15    &   0.17   \\
                                                   & OPD \cite{jiang2022opd}      &    0.22     &     0.25    &   0.19         &   0.21    &     0.25      &   0.27   &   0.19     &   0.18     &  0.22    \\
                                                & Ours - ATOP                    &  \cellcolor{blue!15} \textbf{0.06}      & \cellcolor{blue!15}  \textbf{0.07}      &     0.54       &   0.19    & \cellcolor{blue!15} \textbf{0.12}         &   \cellcolor{blue!15}\textbf{0.09}   &     \cellcolor{blue!15}\textbf{0.11}   &    \cellcolor{blue!15}\textbf{0.04}    &   \cellcolor{blue!15}\textbf{0.16}   \\ \hline
\end{tabular}
\vspace{2mm}
\caption{
Quantitative evaluation of zero-shot generalization on ACD dataset \cite{iliash2024s2o} curated by \cite{liu2024singapo}. Motion parameters are optimized with guidance of personalized motion diffusion model with {\em no new finetuning \/}.
}
\label{tab:motion_parameter_estimation_gen}
\vspace{-5mm}
\end{table*}

\subsection{3D Motion Axis Optimization from Multi-View Motion} 
\label{sec: 3D motion inference}

We leverage the personalized motion as a prior to robustly estimate the 3D motion parameters on a static target 3D mesh. 

\subsubsection{Estimating intermediate states:} Given the input shape which is represented as a mesh $\mathcal{M}$, with vertices $\mathcal{V} \in \mathbb{R}^{n\times3}$ and faces $\mathcal{F} \in \{1, 2, \dots, n\}^{m\times3}$, we deform the mesh using the guidance from the personalized multi-view motion diffusion model. For this, we first parameterize the 3D representation for each frame of the video using $N_{g}$ gaussians which are initialized with the points sampled from the input mesh. This representation is then optimized by distilling the motion priors encapsulated in the personalized frozen diffusion model $(\psi^{*})$ using the score distillation loss (Eq. \ref{eq: sds_actual}). At each optimization step illustrated in Fig. \ref{fig:pipeline-3d-motion-transfer}, we use a differentiable renderer $\mathcal{R(.)}$ \cite{keselman2022approximate, keselman2023flexible} to render multiple views of per-frame gaussians in 3D space to its corresponding frame in latent space of the diffusion model. The motion video rendered by these gaussians is then denoted as $\mathcal{Z}_{g}$. Next, we sample a random noise $\epsilon \thicksim \mathcal{N}(0,1)$ at each diffusion timestep $t$ and add it to the rendered latents $\mathcal{Z}_{g}$ to obtain noisy latents. These noisy latents are then denoised using the personalized multi-view motion model $(\psi^{*})$ (Sec~\ref{sec:ft_for_mv_video-gen})
where the diffusion model is conditioned using the text prompt, $\tau$, rendered multi-view images $\mathcal{I}_{r}$ and segmentation masks $\mathcal{B}$ of the target mesh. This optimization results in a sparse point cloud of the articulated shape for each temporal step. 
\subsubsection{Motion Axis Estimation:} Given this sparse point cloud of intermediate articulated shape, our goal is to then estimate the motion attributes: \textit{motion axis} and \textit{motion origin}. We focus on piecewise rigid motions, specifically \textit{revolute} and \textit{prismatic} which are the most widely appearing motion types in nature \cite{liu2023paris, qiu2025articulate}. Details on how the motion is represented is described in Appendix. Having obtained a sparse point cloud of articulated shape for each time step, we can directly optimize the motion axis parameters using chamfer distance loss. However, it might happen that because of sparse views, the reconstructed point cloud is noisy which affects this chamfer distance based optimization. Hence, to further improve the accuracy we propose an algorithmic approach for the same. In this step, given the static mesh $\mathcal{M}$, we sample points from the segmented part to be articulated and approximate it with an oriented bounding box (OBB). From the OBB parameters, we determine 7 possible axis origins—one centroid and the centers of its six side faces—and 6 possible axis directions, derived from its three principal components and their negations. The dynamic part is then transformed under a piecewise rigid assumption, and we select the best axis and origin by minimizing the chamfer distance with the articulated point cloud. See Appendix for algorithmic details.

\subsection{Training Objectives}
\label{sec: training_objectives}

Our model works in a two-stage setting: (a) Multi-View Motion Generation via Personalization and (b) 3D Motion Axis optimization.
\subsubsection{Multi-View Motion Generation via Personalization} During finetuning, we keep the diffusion model frozen and train only the controllable motion module (Eq. \ref{eq: temp_attention}). The network is optimized using the multi-view diffusion objective $\mathcal{L}_{LDM}$, defined as:
\begin{figure*}[t]
\small
\centering
\includegraphics[scale=0.2]{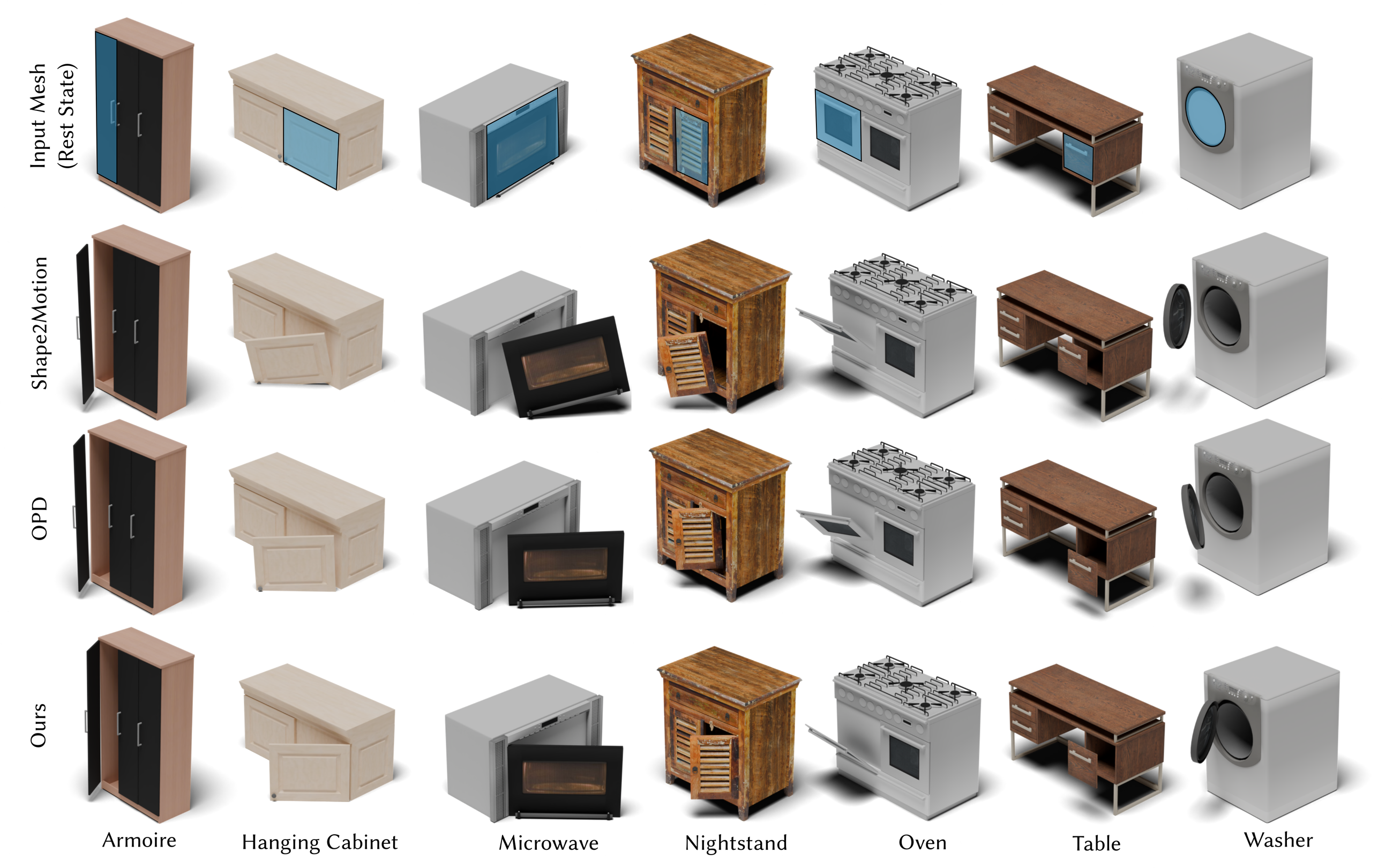}
\caption{Qualitative comparison of generalization capabilities of part articulation on ACD dataset. The part to be articulated is highlighted in first row. As can be seen, our approach can robustly estimate motion parameters in unseen scenarios. Note: The models in the ACD dataset do not have interiors.}
\label{fig:results_acd}

\end{figure*} 
\begin{equation}
\small
\mathcal{L}_{\text{\textit{LDM}}} = \mathbb{E}_{\text{z}, \tau, \mathcal{I}_{\text{r}}, \mathcal{C}, \epsilon, \text{t}} [\|\epsilon - \epsilon_{\theta}(\text{z}_\text{t}; \tau, \mathcal{I}_\text{r}, \mathcal{B}, \mathcal{C}, \text{t})\|^2].
\label{eq:l_ldm}
\end{equation}
where, $\text{z}_{t}$ represents the latents at time step $t$, $\tau$ is the text prompt, $\mathcal{I}_{\text{r}}$ is the reference image, $\mathcal{B}$ is the binary mask, and $\mathcal{C}$ denotes the camera poses.

\subsubsection{3D Motion Parameter Optimization.} During 3D motion parameter optimization, we optimize the gaussians using SDS loss \cite{poole2022dreamfusion}. Since our generation is conditioned on multi-view images $\mathcal{I}_{r}$ and masks $\mathcal{B}$, the loss is defined as:
\begin{equation}
\small
    \nabla_{\theta} \mathcal{L}_{SDS} = \mathbb{E}_{t, \epsilon} \left [ \omega(t)(\epsilon_\phi(\boldsymbol{z}_{g_t}; t, \tau, \mathcal{I}_{r}, \mathcal{B}, \mathcal{C}) - \epsilon) \frac{\partial \boldsymbol{z}_{g_t}}{\partial \theta}\right ]
    \label{eq: sds_actual}
\end{equation}
Here, $\theta$ represents the parameters of the differentiable renderer $\mathcal{R}(.)$, $\epsilon_{\phi}$ is the UNet of the personalized diffusion model $\psi^{*}$, and $z_{g_{t}}$ is the latent rendered by the renderer with noise for timestep $t$. $\tau$, $\mathcal{I}_{r}$, and $\mathcal{B}$ refer to the text prompt, rendered image, and segmentation mask of the target mesh, respectively. $\mathcal{C}$ are the camera poses.
\subsubsection{Training details.} Our implementation is based on the PyTorch framework \cite{paszke2019pytorch}, and for diffusion, we use the widely adopted diffusers library \cite{von-platen-etal-2022-diffusers}. In both stages, fine-tuning and 3D motion parameter optimization, we use the Adam optimizer \cite{kingma2014adam} to update the trainable parameters. During fine-tuning, we set the learning rate to $5 \times 10^{-4}$, and during the 3D motion transfer stage, we use a learning rate of $5 \times 10^{-3}$. The fine-tuning stage is run for 20,000 iterations. Each multi-view video sample consists of 4 views, with 10 frames per view, at a resolution of $256 \times 256$. For the personalization step, we perform inference over 50 steps using a classifier-free guidance scale of 5.0. Both fine-tuning and inference are carried out using the fp16 model weights of ImageDream. We fine-tune a single personalized diffusion model on an NVIDIA V100 GPU with $32$ GB of memory, which takes approximately $2$ hours to complete. At inference time, we sample $2,048$ points each from both the static and dynamic parts of the mesh, and use these to initialize the Gaussian centers. For our renderer, we set $\beta_{1} = 21.4$ and $\beta_{2} = 2.66$. Since optimization is performed in the latent space of the diffusion model, we directly render features from our renderer $\mathcal{R(.)}$ where instead of RGB channels for each gaussian we learn a $4D$ channel feature for each gaussian. We then render features at a resolution of $32 \times 32$ which is equal to the latent space dimension of the diffusion model.

\section{Experiments}
\label{experiments}
\newcommand{\TODO}[1]{\textbf{\color{red}[TODO: #1]}}

\begin{figure*}[t]
\centering
    \includegraphics[width=\textwidth]{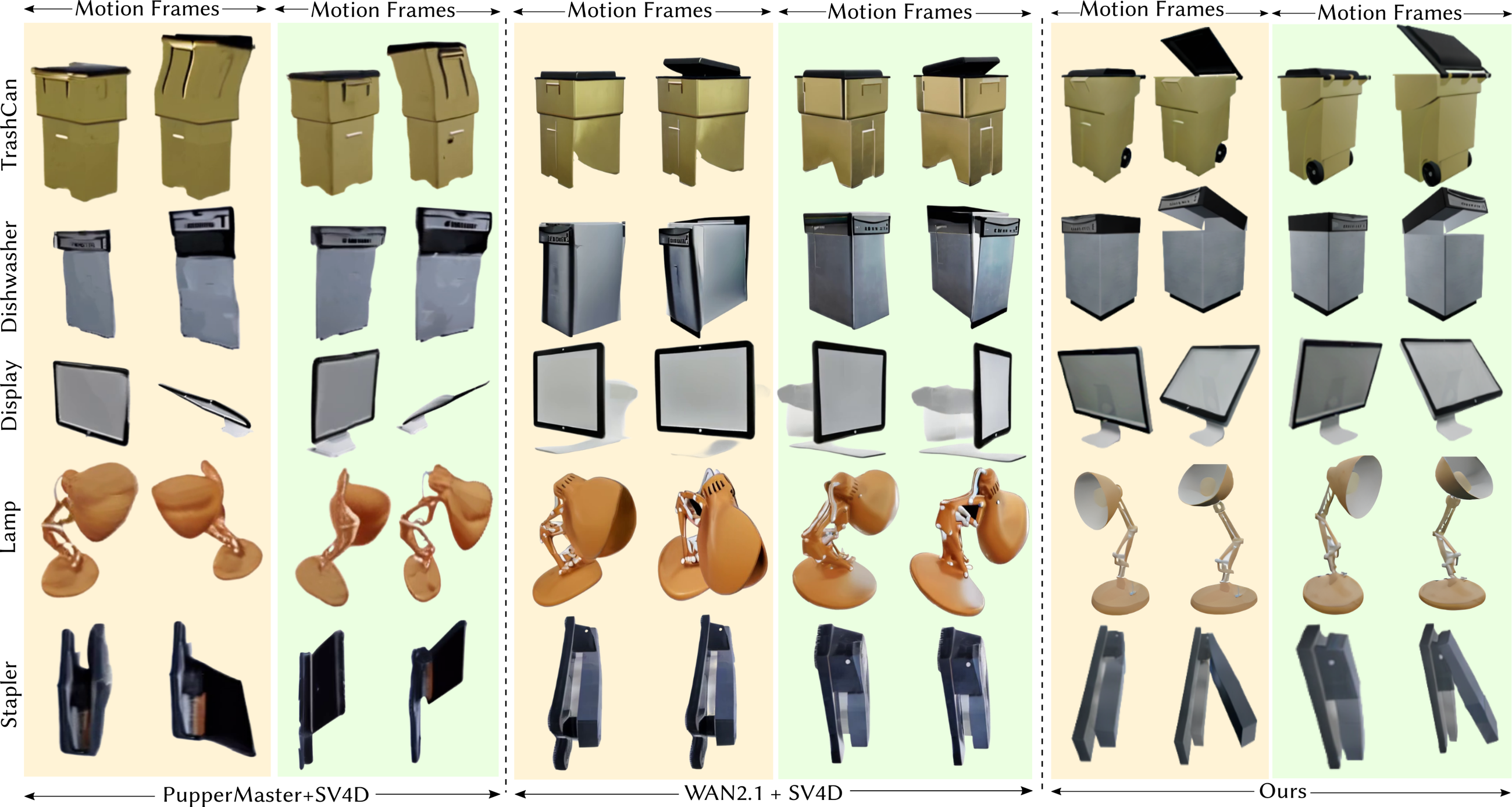}
\caption{Qualitative comparison of our method with proposed Multi-View motion generation baselines on Objaverse and PartNet-Mobility objects. Different color columns indicate different views. As can be seen our approach can generate more realistic and plausible part motion videos when compared against more general baselines which are trained on large scale datasets for a {\em general task \/}, showcasing the effectiveness of our few-shot approach for this task.}
\label{fig:results_sv4d}
\end{figure*}
\begin{figure*}[t]
\centering
    \includegraphics[width=\textwidth]{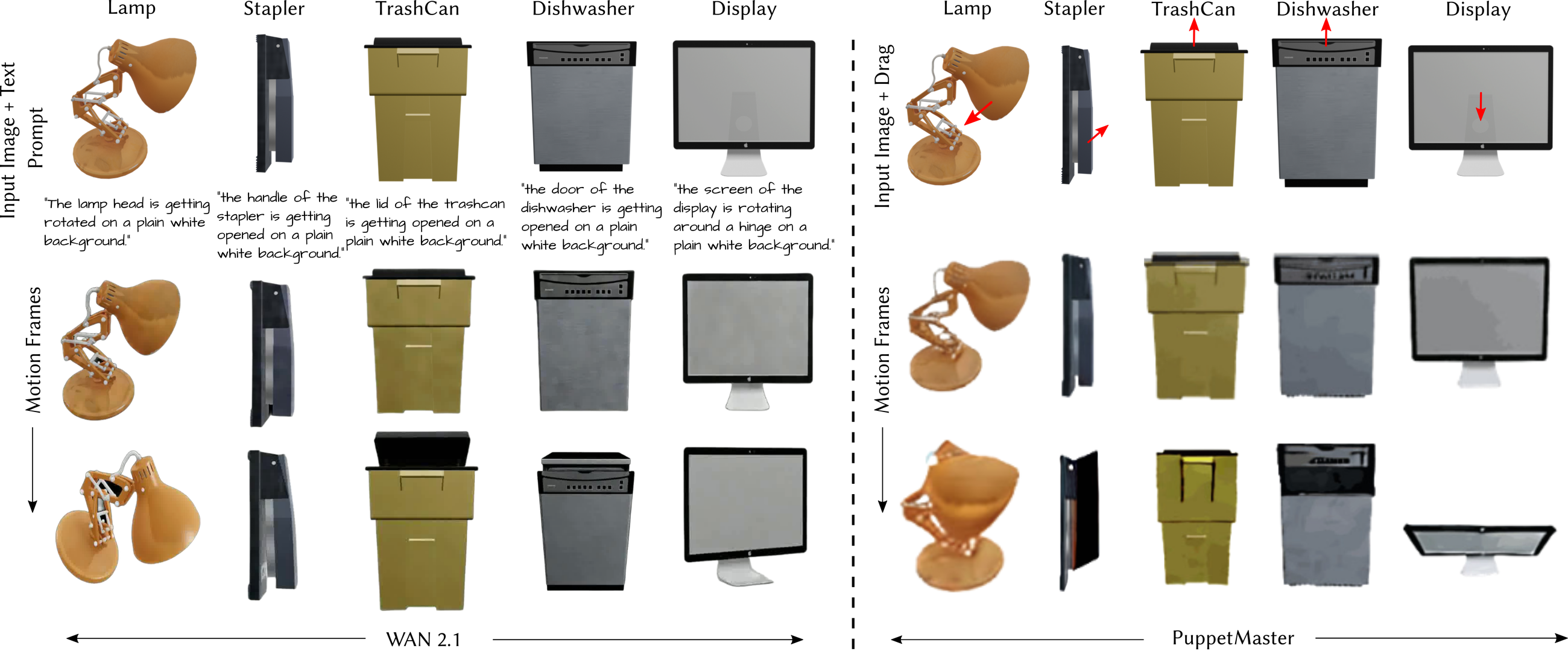}
\caption{Qualitative comparison results of I2V baselines. Left is the results generated by an I2V model WAN 2.1 \cite{wan2025wan} which generates a video from a text prompt given an image as input. Right are the results generated from an another I2V model PuppetMaster \cite{li2025puppet} which takes ``drag" as a prompt. As can be seen, none of the prompting methods yields better and controllable video output. Since ours is a single stage pipeline which directly predicts multi-view motion output, we skip this comparison and directly compare the multi-view output as shown in Fig. ~\ref{fig:results_sv4d}.}

\label{fig:results_mv}
\end{figure*}

In this section, we present qualitative and quantitative results demonstrating our method's ability to generate high quality multi-view video outputs and, using this capacity, infer 3D motion parameters given only a static segmented mesh.

\subsection{Dataset}
We train our approach on the PartNet-Mobility dataset \cite{xiang2020sapien}, following the train-test split of \cite{kim2024partstad, liu2023partslip}:
for each category, $8$ shapes are used for training, and the rest for testing. 
For each articulated part in a training shape, we render a four-view video depicting the articulation.
Views are set at azimuth angles of $[45\degree, 135\degree, 225\degree, 315\degree]$, with elevation at $10\degree$ for some categories and $30\degree$ for rest. 
These views are selected based on how closely they match the training distribution of camera poses of ImageDream~\cite{wang2023imagedream}, and fixed throughout our experiments. 
Each view consists of $10$ frames at $256\times256$ resolution, along with the first frame's binary mask of the part that is to be articulated. 
To evaluate the generalization capacity of our method, we also conduct {\em zero-shot \/} experiments on $135$ objects from a subset~\cite{liu2024singapo} selected from the ACD dataset~\cite{iliash2024s2o}, which features challenging objects from ABO~\cite{collins2022abo}, 3D-Future~\cite{fu20213d} and HSSD~\cite{khanna2024habitat} that are structurally distinct from each other. ACD provides textured segmented meshes with part annotations. We render multi-view images and segmentation masks using the same camera viewpoints used during finetuning. To further evaluate the generalization ability of our method, we test it on the Objaverse dataset \cite{deitke2023objaverse, deitke2024objaverse}. We curate a small subset of objects by selecting those that appear among the top-retrieved results when queried with relevant text prompts corresponding to the target category. Since, objaverse do not have any segmentation annotations, we run off-the-shelf open-vocabulary segmentation method like PartSTAD ~\cite{kim2024partstad} to get the required segmentation. We then render multi-view images and segmentation masks of these objects in the same way as done with PartNet-Mobility ~\cite{xiang2020sapien} and ACD ~\cite{iliash2024s2o} datasets. We then infer multi-view motion output from these multi-view images and masks as input to our DM. Since Objaverse dataset does not have ground-truth motion attributes annotations, we only show qualitative results on the collected samples to demonstrate the generalization ability of our method.
\begin{figure}[t]
\small
\centering
\includegraphics[scale=0.125]{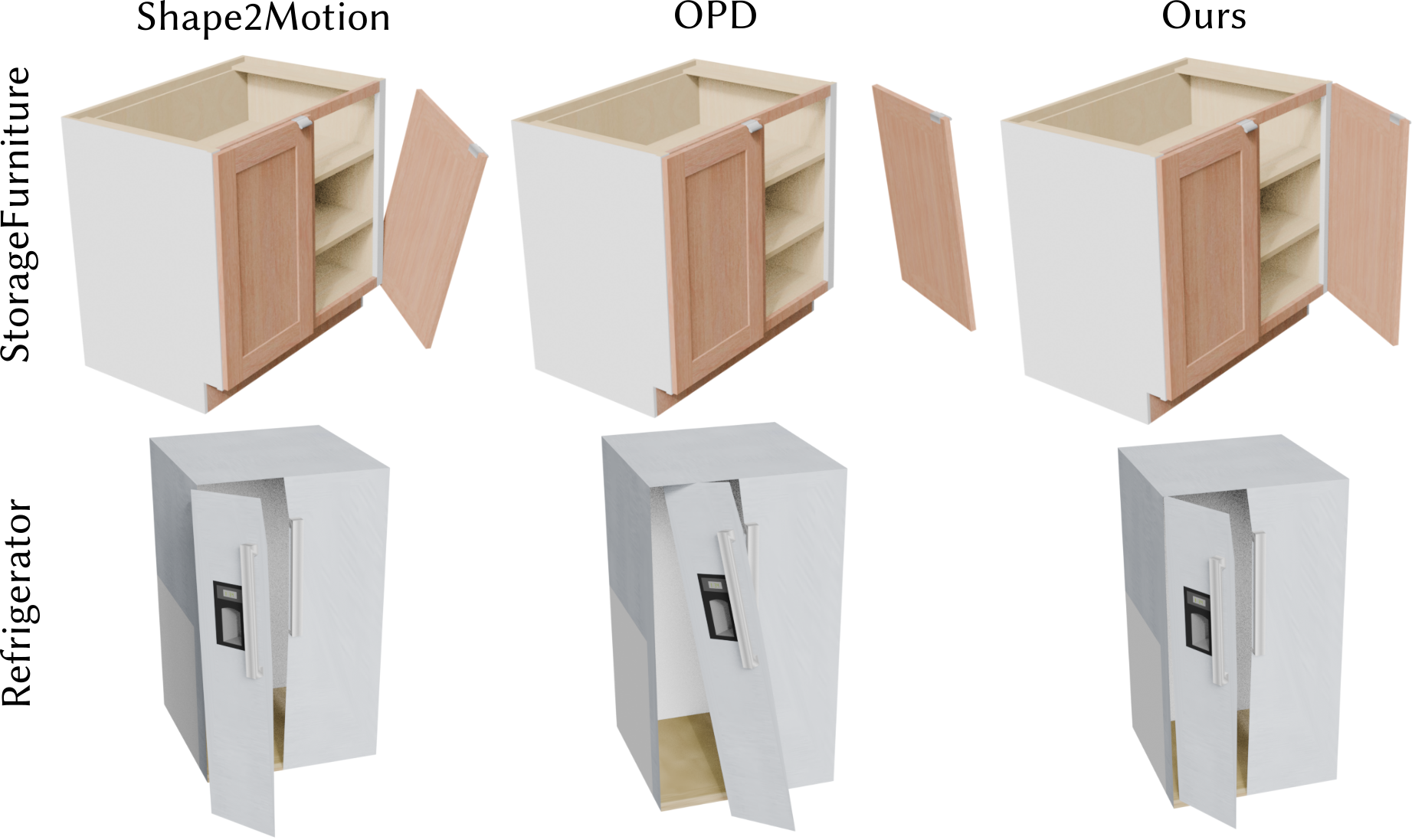}
\caption{Qualitative results of part articulation on the PartNet-Mobility dataset. Note that our approach yields most plausible motion parameter estimation when compared to the baselines.}
\label{fig:results_ps}
\vspace{-5mm}
\end{figure}
\begin{table}[t]
\centering
\small 
\begin{tabular}{c|c|c|c|c}
\toprule
\textbf{Method} & \texttt{CLIP} $(\uparrow)$ & \texttt{FVD} $(\downarrow)$ & \texttt{LPIPS} $(\downarrow)$ & \texttt{PSNR} $(\uparrow)$ \\
\midrule
WAN + SV4D & $26.21$ & $1456.83$ & $0.569$ & $11.38$ \\
PM + SV4D & $24.36$ & $1327.32$ & $0.487$ & $12.62$ \\ \hline
Ours & \cellcolor{blue!15}\textbf{26.47} & \cellcolor{blue!15}\textbf{938.97} & \cellcolor{blue!15}\textbf{0.128} & \cellcolor{blue!15}\textbf{20.87} \\
\bottomrule
\end{tabular}
\vspace{2mm}
\caption{Quantitative evaluation of video generation}
\label{tab:video_generation_comparison}
\end{table}

\begin{table}
    \begin{minipage}{.5\linewidth}
          \centering
        
        \label{tab:runtime}
        \vspace{-0.1in}
        \begin{tabular}{c|cc}
\hline
Method       & \texttt{MAE} $(\downarrow)$          & \texttt{MPE} $(\downarrow)$                   \\ \hline
Dir. Opt.        & $1.57$          & $0.11$           \\
Algo. Opt. & \cellcolor{blue!15}\textbf{0.98} & \cellcolor{blue!15}\textbf{0.07} \\ \hline
\end{tabular}
\vspace{2mm}
\caption{Dir. Opt. vs. Algo. Opt.}
\label{tab:dir_vs_algo.}
    \end{minipage}%
    \begin{minipage}{.5\linewidth}
 
 \label{tab:user}
      \vspace{-0.1in}
      \small
      \centering
\begin{tabular}{c|cc}
\hline
\# Vids & \texttt{FVD} $(\downarrow)$ & \texttt{PSNR} $(\uparrow)$ \\ \hline
1    & 1296.2            &  14.23     \\
4  & 1172.3             & 17.78   \\ \hline

\cellcolor{blue!15}  8   & \cellcolor{blue!15}\bf 983.7 & \cellcolor{blue!15}\textbf{19.83}                \\ \hline
\end{tabular}
\vspace{2mm}
\caption{Training number of videos.}
\label{tab:num_vids}
    \end{minipage}
    \vspace{-0.3in}
\end{table}

\subsection{Multi-View Motion Generation}
We perform qualitative and quantitative evaluation of our multi-view generation output on a variety of classes from PartNet-Mobility dataset ~\cite{xiang2020sapien} and Objaverse dataset ~\cite{deitke2023objaverse} like {\em trashcan, dishwasher, display, lamp, storage and stapler\/}. For quantitative evaluation we only consider objects from PartNet-Mobility dataset as the Objaverse dataset does not have ground truth multi-view motion. Figure~\ref{fig:teaser}, middle column shows non cherry-picked qualitative examples of multi-view motion frames generated by our method on Objaverse dataset objects. More qualitative results on variety of test shapes from ACD~\cite{iliash2024s2o}, Objaverse~\cite{deitke2023objaverse, deitke2024objaverse} and PartNet-Mobility ~\cite{xiang2020sapien} is shown in Fig. ~\ref{fig:acd_new_1}, ~\ref{fig:acd_new_2}, ~\ref{fig:objaverse_new_3}, ~\ref{fig:ps_new_4} in the supplementary material. As shown in the figures, our method is capable of generating plausible and spatio-temporally consistent multi-view motion outputs from input multi-view images, while also preserving piecewise rigidity—an essential property for accurately transferring articulation motion from 2D to 3D space. Additionally, our approach supports {\em spatial control\/} of the generated motion using binary masks, enabling articulation of specific parts of a shape.

Since no prior work directly predicts consistent {\em multi-view motion} from {\em static multi-view images \/} in a {\em controllable manner\/}, we create two-step baselines that first generate single videos from a static image, and then expand the single-video video to a set of multi-view videos.
In the first step, we use two off-the-shelf state-of-the-art I2V diffusion models with varying forms of controls:
\begin{enumerate}[itemsep=2pt, parsep=0pt, topsep=2pt, leftmargin=*]
\item{\em WAN2.1\/}~\cite{wan2025wan} is a powerful video foundation model based on the diffusion transformer framework. It uses a spatio-temporal variational encoder (VAE) to capture both spatial and temporal aspects of videos and is trained on a large-scale dataset of billions of images and videos. WAN $2.1$ is available in two model sizes: $1.3B$ and $14B$ parameters, targeting efficiency and performance, respectively. We use the $14B$ variant for our experiments. All our results for image to video generation are generated using the HuggingFace demo, which offers a distilled version of WAN $2.1$ optimized for fast inference in just $4$ steps using CausVid LoRA. 
\item {\em PuppetMaster\/}~\cite{li2025puppet} is a generalized I2V model which is trained with more explicit prompt controls like ``drag", which are more robust than general text prompts, as observed in our experiments. These drags are manually annotated for a large collection of dynamic content from Objaverse. Once trained, PuppetMaster diffusion model can synthesize a video depicting part-level motion faithful to the given drag interactions. 
\end{enumerate}
\begin{figure*}[t]
\centering
   \includegraphics[width=\textwidth]{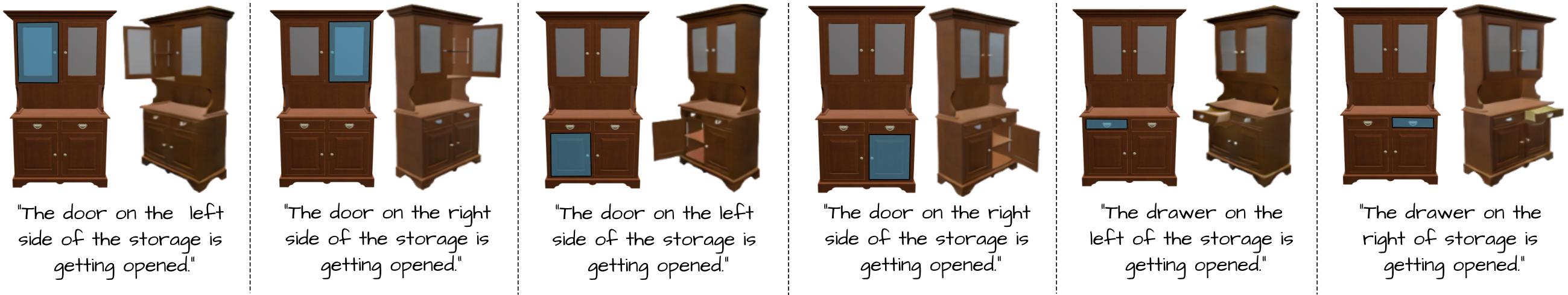}
    \caption{{\em Effect of Masking in Articulation.\/} By providing appropriate part mask as prompt, we can control the spatial location of motion in our diffusion model.}
\label{fig:mask_effect}
\end{figure*}
We then feed the single-view generated by the respective methods into SV4D~\cite{xie2024sv4d, yao2025sv4d}, which is a latent video diffusion model designed for generating {\em multi-view\/} and {\em multi-frame\/} consistent dynamic 3D content from a single-view input video which enables efficient 4D asset optimization. 
(Note that SV4D is trained on a {\em large corpus of dynamic scenes\/} from Objaverse \cite{deitke2023objaverse}, unlike ours which is trained on only {\em few-shot\/} reference motion samples.)
We denote these baselines \text{WAN2.1+SV4D\/} and \text{PM+SV4D\/} respectively.

As our method is a single-stage pipeline that directly predicts multi-view motion outputs, we compare multi-view outputs and skip I2V comparison. 
We report performance on standard metrics that measure video quality: 
Frechet Video Distance (\texttt{FVD}) \cite{unterthiner2019fvd}, which measures temporal coherence;
\texttt{CLIP} score \cite{CLIP1}, which measures text alignment;
\texttt{PSNR} and \texttt{LPIPS}~\cite{zhang2018perceptual}, which measures spatial and temporal consistency. These metrics are computed only for objects of PartNet-Mobility dataset, as the Objaverse dataset does not provide ground-truth multi-view motion.

Table~\ref{tab:video_generation_comparison} summarizes the results. 
Our method significantly outperforms the baselines across all quantitative metrics, particularly in terms of \texttt{PSNR} and \texttt{LPIPS}, highlighting the spatial and temporal consistency of the generated multi-view video output with respect to the ground truth. These results demonstrate the effectiveness of our personalization step in producing outputs that are piecewise rigid and faithfully aligned with the input target shape.

We further show qualitative comparisons in Figure~\ref{fig:results_sv4d}.
While all methods generate $4$ views, we only show two views for simplicity, with a different background color for each view.
Our method is able to produce more realistic and plausible part motion videos compared to the baselines, which struggles due to combination of failure modes in each of the steps:
\begin{enumerate}[itemsep=2pt, parsep=0pt, topsep=2pt, leftmargin=*]
\item The I2V diffusion models cannot consistently create high-quality motions. We show the intermediate output of these models in Figure~\ref{fig:results_mv}. Despite being a large model trained on a large-scale dataset, WAN2.1 lacks a clear understanding of part-specific dynamics. This results in incorrect outputs in several cases when motion is controlled using only a text prompt. For example, for the {\em lamp\/} and {\em display\/} categories, the model generates videos where the entire lamp rotates upside down instead of just the lamp head. In the case of the {\em dishwasher\/}, the model fails to understand how the door opens based on the input and instead produces a motion pattern it may have encountered during training. Puppetmaster, while having more explicit controls, often struggle to disambiguate between parts using only drag prompts, leading to incorrect motion and poor quality video.
\item SV4D also introduces new geometric inconsistencies, especially when it comes to motions. This can be especially observed in the output of trashcan and dishwasher of Fig. ~\ref{fig:results_sv4d} where SV4D is not able to faithfully reconstruct the dynamic content in its multi-view motion output. 
\end{enumerate}

\subsection{3D Motion Axis Parameter Estimation}
We perform qualitative and quantitative evaluation of our 3D motion axis parameter estimation output on a variety of classes from PartNet-Mobility ~\cite{xiang2020sapien}, ACD ~\cite{iliash2024s2o} and Objaverse datasets ~\cite{deitke2023objaverse}. For quantitative evaluation we only consider objects from categories of PartNet-Mobility and ACD dataset as the Objaverse dataset does not have ground truth 3D motion axis annotations. Figure~\ref{fig:teaser}, right column shows non cherry-picked examples of motion parameters predicted by our method from multi-view videos, generated from views rendered from a static, segmented 3D mesh.
We also show the predicted 3D motion along with the aforementioned multi-view video results in Fig.~\ref{fig:acd_new_1}, ~\ref{fig:acd_new_2}, ~\ref{fig:objaverse_new_3}, ~\ref{fig:ps_new_4}  in the supplementary material. As shown in these qualitative results, our method successfully transfers the multi-view motion from 2D back to 3D, demonstrating strong spatial consistency in the generated output. This consistency is crucial for effectively optimizing the motion axis parameters. Furthermore, our algorithmic optimization approach enhances accuracy by constraining the search space to a fixed set of possible solutions, making the estimation process more efficient and reliable.

We compare motion axis parameter estimation step against two representative baselines for motion prediction:
\begin{enumerate}[itemsep=2pt, parsep=0pt, topsep=2pt, leftmargin=*]
    \item Shape2Motion (S2M)~\cite{wang2019shape2motion} 
    is a method that predicts motion parameters from a point cloud and is trained end-to-end with 3D ground truth motion axis parameters. 
    \item OPD~\cite{jiang2022opd}
    is a method that estimates motion parameters from a single image using an object detector backbone that predicts segmentation masks, motion type, and motion parameters. 
\end{enumerate}
We compare against both baselines in a few-shot setting by retraining them with our train and test split. To assess 3D motion axis parameter accuracy, we follow these baselines and report 1) Mean Angular Error (\texttt{MAE}), which is the dot product between predicted and ground truth motion axes; 2) Mean Position Error (\texttt{MPE}), which is the euclidean distance between their joint origins.

Table~\ref{tab: partnet_sapian_quant} and Figure~\ref{fig:results_ps} summarizes the results on PartNet-Mobility.
Our method predicts plausible motion parameters without using 3D annotations, beating the prior state-of-the-art methods by a large margin, especially in terms of angular error (\texttt{MAE}).
We further evaluate generalization performance in Table~\ref{tab:motion_parameter_estimation_gen} and Figure~\ref{fig:results_acd}.
Shape2Motion, which relies on explicit 3D point clouds, struggles with generalization, evident from its relative performance against OPD.
OPD, which relies on {\em single-image\/} inputs, generalize much better, but still struggles with raw performance due to lack of 3D awareness.
In contrast, our method generalizes well, showcasing the effectiveness of the diffusion prior in learning part dynamics and the 3D motion axis parameter estimation step which reliably lifts the motion from 2D to 3D space. 
Additionally, our algorithmic optimization strategy simplifies the motion axis estimation step by transforming it from a continuous regression task into a discrete classification task. 
This reformulation allows us to search for the optimal solution within a fixed solution space, making the optimization process more tractable compared to the regression-based approaches used in prior works. 
As a result, we are able to achieve more accurate outcomes for 3D motion axis parameter estimation. As shown in Fig.\ref{fig:results_acd}, objects such as {\em Armoire}, {\em Nightstand}, and {\em Table} exhibit significant structural differences from the training video samples presented in Fig.\ref{fig:training_samples}. Despite these differences, our model is able to faithfully transfer the motion from 2D space back to 3D, demonstrating its strong generalization capability.
\begin{figure}[t]
\centering
   \includegraphics[width=\columnwidth]{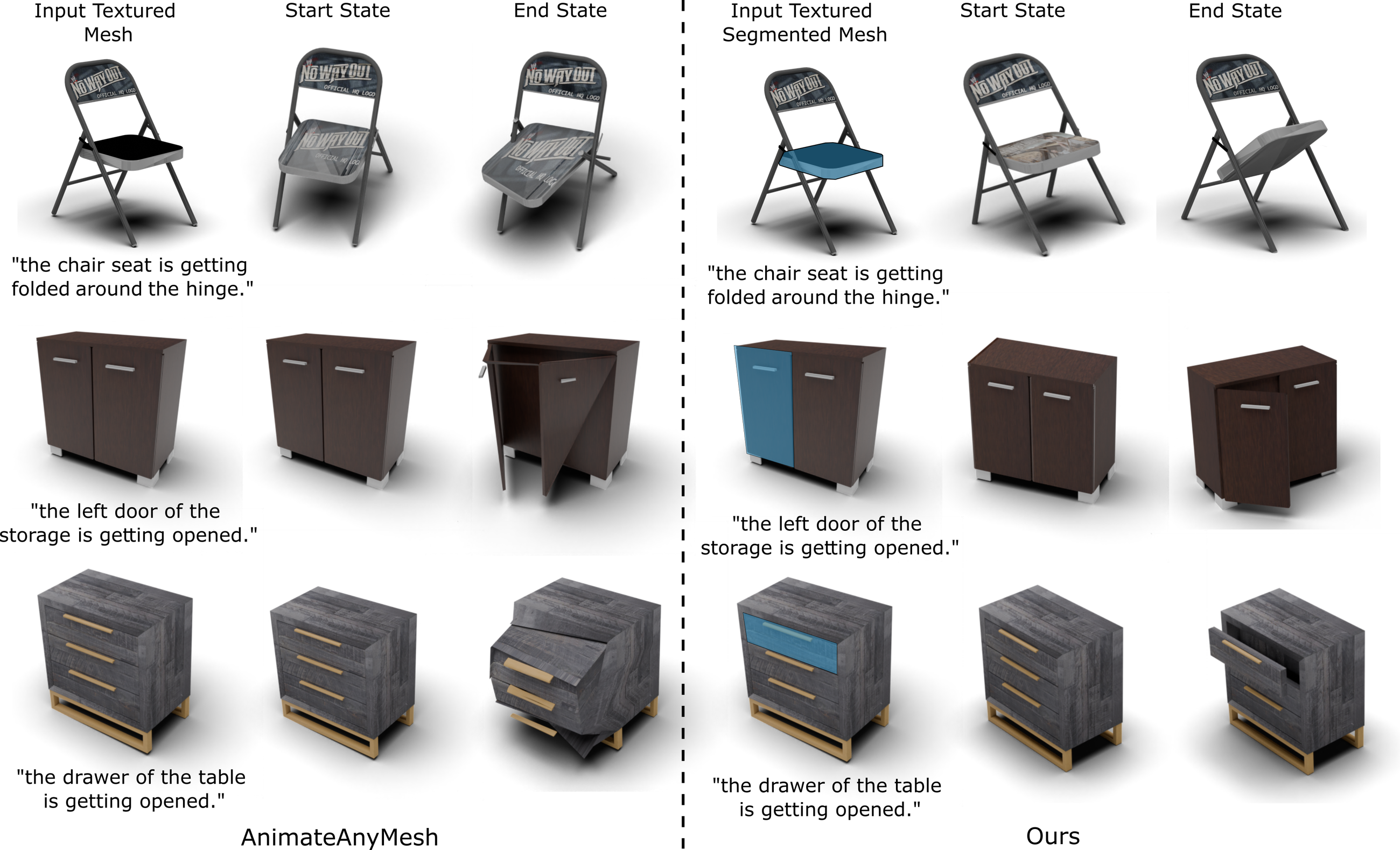}
    \caption{\rev{{\em Rigid vs. Soft Deformation:} We compare our {\em rigid-constrained} based articulation method (right) with {\em soft deformation} based mesh animation approaches (left) \cite{wu2025animateanymesh}. Our method produces more accurate and physically plausible animations, whereas unconstrained soft deformation methods often lead to irregular deformations.}}
\label{fig:rigid_vs_soft}
\vspace{-5mm}
\end{figure}

\begin{figure}
\centering
   \includegraphics[scale=0.15]{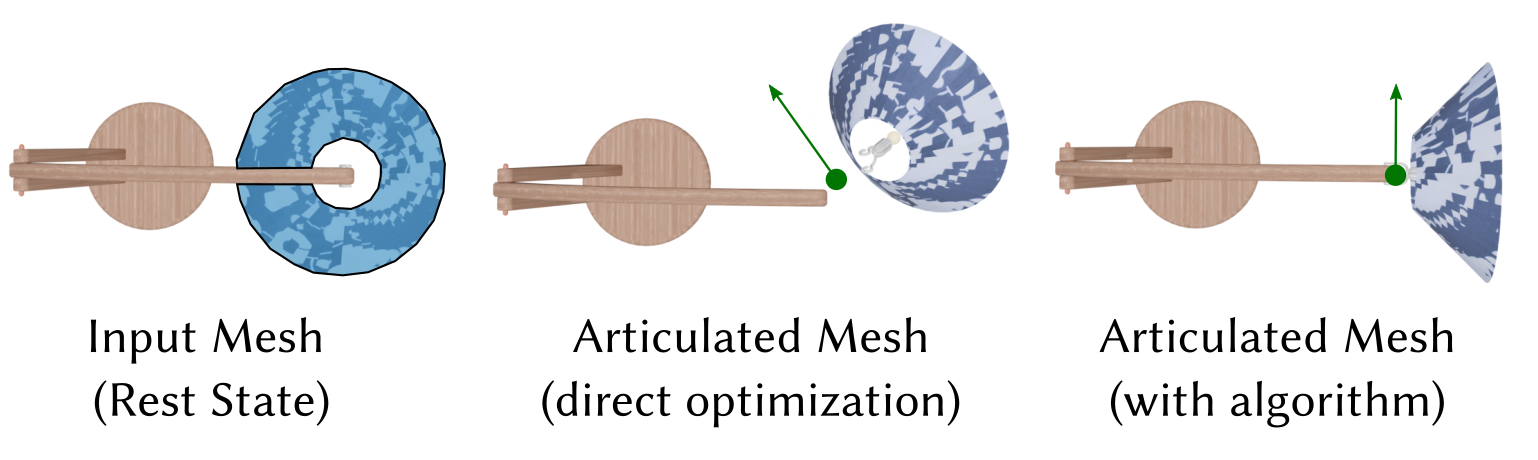}
    \caption{Motion axis origin and orientation estimation using direct and algorithmic optimization. Because of restricting the solution space using OBB, we can estimate more plausible solution (right) compared to direct optimization (middle). Part to be articulated is highlighted (left).}
\label{fig:dir_vs_algo}
\vspace{-5mm}
\end{figure}

\section{Ablations} 
\label{sec:ablations}
\begin{figure}[t]
\centering
   \includegraphics[scale=0.275]{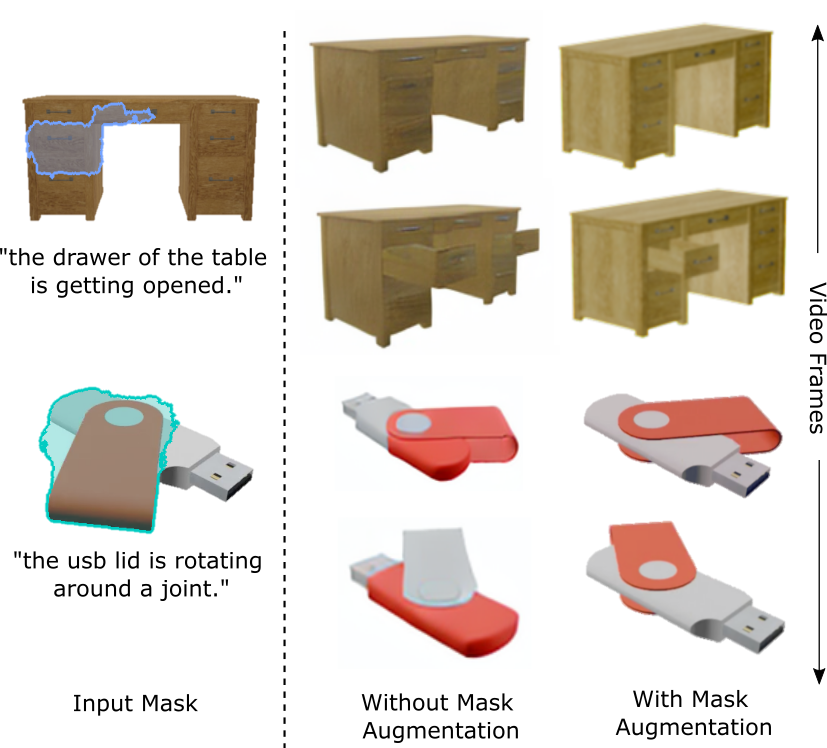}
    \caption{\rev{{\em Effect of mask quality in motion generation:} We observe that noisy mask predictions from foundation models such as SAM \cite{ravi2024sam2segmentimages} can impact the motion generation module (middle column), particularly its spatial control. However, this issue can be effectively mitigated through mask-based data augmentation during training (right column).}}
\label{fig:err_mask_effect}
\vspace{-5mm}
\end{figure}

\begin{figure}[t]
\centering
   \includegraphics[scale=0.25]{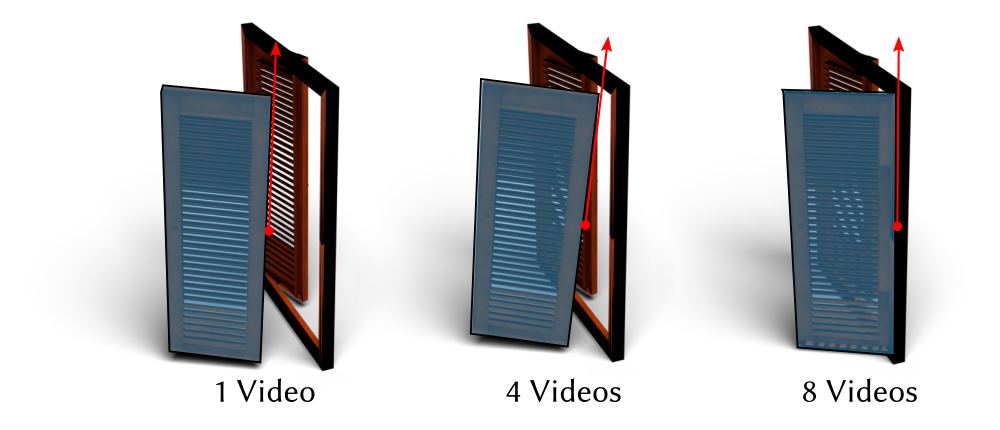}
    \caption{Effect on motion axis estimation using guidance from a diffusion model trained on increasing number of videos (left to right). As the number of videos used for training are increased, more accurate articulation attributes can be estimated.}
\label{fig:num_videos}
\vspace{-5mm}
\end{figure}

\begin{figure*}[t]
\centering
   \includegraphics[width=\textwidth]{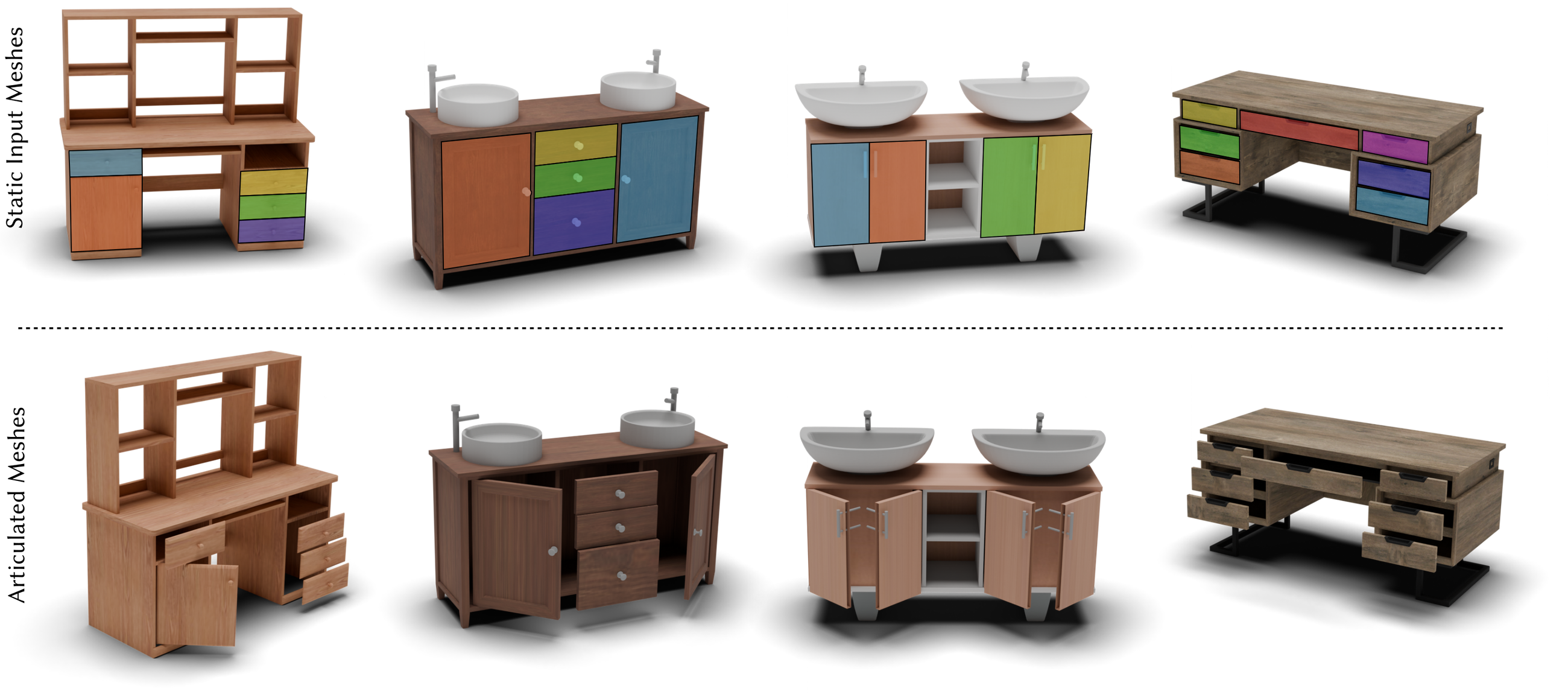}
    \caption{{\em Multi-part articulation:\/} We demonstrate that by running inference multiple times with our personalized diffusion model, each time modifying the spatial location of the mask (indicated by different colors for each object), we can successfully articulate multiple parts of the same object.}
\label{fig:multi-part}
\end{figure*}

\begin{figure*}[t]
\centering
   \includegraphics[width=\textwidth]{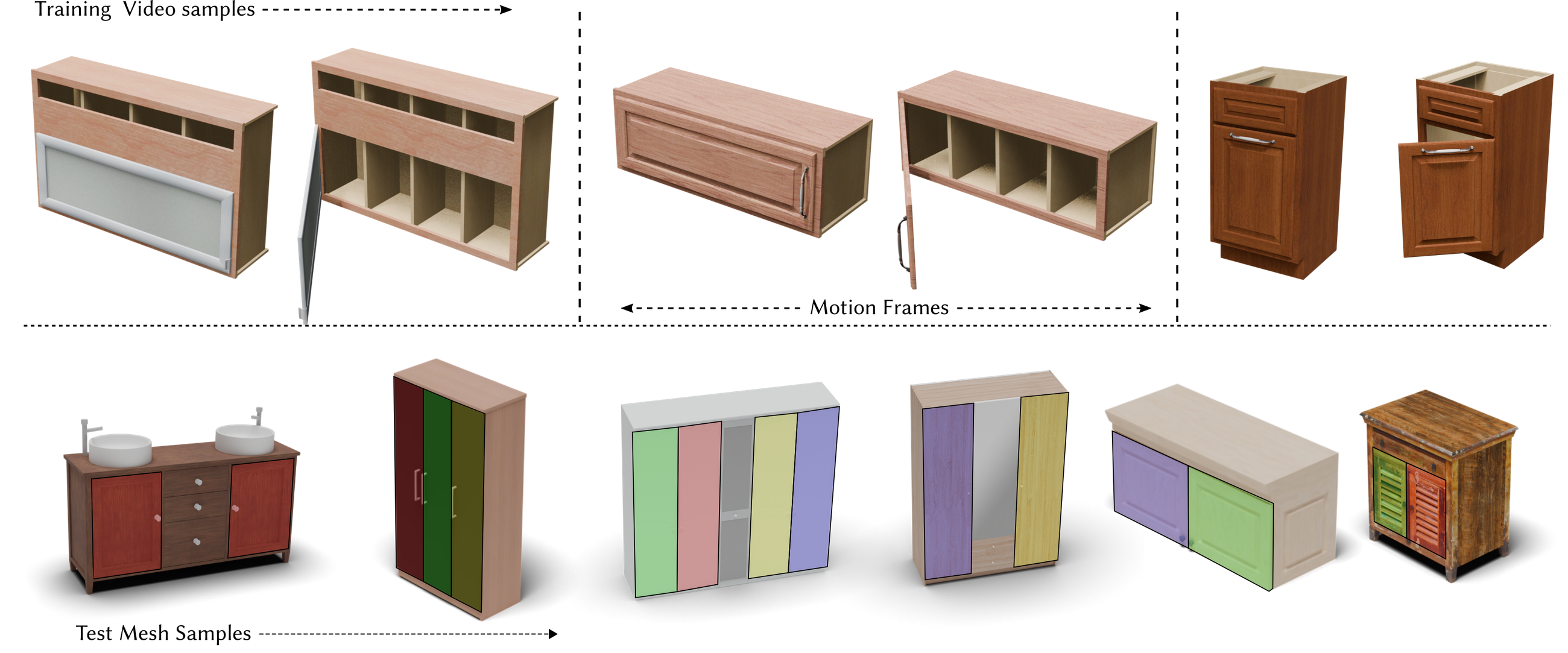}
    \caption{Top row shows video samples used for finetuning the diffusion model. The samples are from PartNet-Mobility dataset ~\cite{xiang2020sapien} following the train split of \cite{kim2024partstad} for StorageFurniture category. The bottom row is the few samples of test meshes on which we estimate the articulation parameters. }
\label{fig:training_samples}
\end{figure*}

We perform various ablations of our method to show the effectivenes of each component in our pipeline. 
First, we perform an ablation on how our algorithm based optimization yields more plausible solution compared directly optimizing the points in the rest state using chamfer distance loss. 
Table ~\ref{tab:dir_vs_algo.} and Fig. ~\ref{fig:dir_vs_algo} shows the quantitative and qualitative results of direct vs algorithmic optimization. 
As can be seen in Fig. ~\ref{fig:dir_vs_algo}, optimizing the lamp head directly with chamfer distance loss yields incorrect output because of noisy point cloud initialization obtained after reconstruction. 
Because we reformulate a continuous regression-based task of motion axis estimation into a discrete classification task, it makes the optimization process more tractable; hence, we are able to achieve more accurate outcome as seen in the Fig. ~\ref{fig:dir_vs_algo} (right). \rev{We further compare our rigidity-constrained articulation method with recent \emph{soft mesh deformation} approaches, such as AnimateAnyMesh \cite{wu2025animateanymesh}, to demonstrate the benefits of rigid constraints. As shown in Fig. \ref{fig:rigid_vs_soft}, soft per-vertex deformation often leads to irregular artifacts, such as shearing in the first row where the seat stretches non-uniformly. A similar issue appears in the third row, where the drawer vertices deform inconsistently during articulation.}
In Table \ref{tab:num_vids} we show how the generation quality of multi-view videos is affected as we increase the number of videos for finetuning. As can be seen in Table \ref{tab:num_vids} multi-view and temporal consistency increases as we increase the number of training videos, which is reflected in the metrics, especially in \texttt{PSNR} which evaluates pixel-level correspondences between the generated output and ground truth. In support to this, in Fig. ~\ref{fig:num_videos} we show the effect on final motion axis estimation using the guidance of different diffusion models trained with increasing number of videos. As can be seen in Table ~\ref{tab:num_vids} and Fig. ~\ref{fig:num_videos}, increasing the number of videos for training yields better geometrically consistent videos which in general yields better estimation of the motion axis.
Finally, in Fig. \ref{fig:mask_effect} we show the motion controllability of our method. As can be seen, by providing the appropriate part mask as prompt, we can control the spatial location of part motion for same object. The spatial controllability of motion plays a key role in enabling the articulation of different parts of the same object. By simply modifying the mask used as a prompt, we can generate appropriate motion for the desired part of the shape. 
This is illustrated in Fig.~\ref{fig:multi-part}, where we present multi-part 3D articulation results across various 3D shapes. 
In each case, we adjust the prompt location based on the specific part to be articulated and then optimize the 3D motion parameters using guidance from the generated multi-view motion output. 
Multiple inference runs are performed by changing the mask location to produce distinct part-specific motions. 
Finally, in Fig.~\ref{fig:training_samples}, we present the training samples used during the finetuning of our diffusion model for one of the category from PartNet-Mobility dataset (StorageFurniture), along with the test shapes from the ACD dataset. 
As shown in the figure, there are substantial structural differences between the training and test objects, highlighting the strong generalization ability of our method.
\rev{In Fig. \ref{fig:err_mask_effect}, we analyze the impact of erroneous masks on motion generation from the personalized diffusion model. When the model is fine-tuned using only clean masks but is given noisy masks at inference time, the spatial control module is negatively affected, leading to inaccurate motion predictions. As shown in the middle column of Fig. \ref{fig:err_mask_effect}, the model may predict motion at incorrect spatial locations or generate incorrect colors, which subsequently affects the final articulation prediction. This issue can be effectively mitigated by augmenting the training data with erroneous masks, enabling the model to become robust to such noise, as illustrated in the right column of Fig. \ref{fig:err_mask_effect}.}



\section{Limitations}
\label{sec:limitations}

While ATOP achieves high accuracy in 3D part articulation, it has some limitations. First, our method currently supports only revolute and prismatic motions and does not handle multiple moving parts simultaneously, though it can infer different part motions in separate inference steps as shown in Fig. ~\ref{fig:multi-part}. Second, our model requires a segmented mesh as input, which can sometimes be a constraint. However, with recent advances in vision-language models (VLMs)  \cite{ravi2024sam2segmentimages, liu2023partslip, kim2024partstad} and diffusion \cite{perla2025asia} based 3D mesh segmentation methods, obtaining segmented meshes has become more accessible. By rendering multiple views of a mesh, VLMs can generate part segmentation, which can then be mapped back to the original mesh. 

\rev{\noindent \textbf{Kinematic Generalization:} We note that kinematic generalization, i.e., inferring new motion types, requires extrapolating motion dynamics beyond the support of the few-shot fine-tuning data. Achieving this would require learning additional priors or “rules” such as explicit joint concepts, physical constraints, or motion-centric representations. In contrast, the proposed method is designed to support geometric generalization, i.e., applying known motion patterns observed during few-shot fine-tuning to new shapes with significant structural variations. Fig. ~\ref{fig:training_samples} in the main paper provides examples of the types of structural variations our model can handle. This design choice is motivated by the limited availability of articulated motion datasets and the strong ability of diffusion models to represent object geometry. By learning a motion pattern under limited supervision, the model can effectively transfer this pattern to novel shapes with diverse structures. 

\noindent \textbf{Independent multi-part motion vs. hierarchical motion:} While our method supports multi-part motion, where different parts move independently with respect to a static reference frame, it does not explicitly model hierarchical motion or kinematic chains (e.g., parent–child dependencies). Modeling such structure would require jointly inferring the object’s kinematic tree and estimating relative articulation transformations and 3D motion parameters, which remains an open and challenging research problem—especially when only sparse motion signals are available from diffusion models. Since our goal is to study part-aware motion transfer in a few-shot setting, we deliberately avoid imposing explicit kinematic constraints and instead focus on learning consistent and transferable part-level motions. Extending the framework to incorporate hierarchical kinematics is a promising direction for future work.
}

\section{Conclusion and future work}
\label{sec:future}
We introduce ATOP, the first method for adding realistic part articulation to existing 3D assets using text prompts and motion personalization. Unlike works such as \cite{liu2024cage, lei2023nap}, which focus on generating articulated assets, our goal is to enhance static 3D models by inferring motion attributes. In future, we will focus on expanding scalability across object categories and enabling cross-category motion transfer—e.g., opening vehicle and room doors using few-shot learning from cabinet-opening videos. More broadly, our approach is a step toward animating vast existing 3D assets, but challenges like the "missing interior" problem remain, as most models lack interiors.

\clearpage

\bibliographystyle{ACM-Reference-Format}
\bibliography{main}


\begin{thebibliography}{85}


\ifx \showCODEN    \undefined \def \showCODEN     #1{\unskip}     \fi
\ifx \showISBNx    \undefined \def \showISBNx     #1{\unskip}     \fi
\ifx \showISBNxiii \undefined \def \showISBNxiii  #1{\unskip}     \fi
\ifx \showISSN     \undefined \def \showISSN      #1{\unskip}     \fi
\ifx \showLCCN     \undefined \def \showLCCN      #1{\unskip}     \fi
\ifx \shownote     \undefined \def \shownote      #1{#1}          \fi
\ifx \showarticletitle \undefined \def \showarticletitle #1{#1}   \fi
\ifx \showURL      \undefined \def \showURL       {\relax}        \fi
\providecommand\bibfield[2]{#2}
\providecommand\bibinfo[2]{#2}
\providecommand\natexlab[1]{#1}
\providecommand\showeprint[2][]{arXiv:#2}

\bibitem[Abbatematteo et~al\mbox{.}(2019)]%
        {abbatematteo2019learning}
\bibfield{author}{\bibinfo{person}{Ben Abbatematteo}, \bibinfo{person}{Stefanie Tellex}, {and} \bibinfo{person}{George Konidaris}.} \bibinfo{year}{2019}\natexlab{}.
\newblock \showarticletitle{Learning to generalize kinematic models to novel objects}. In \bibinfo{booktitle}{\emph{Proceedings of the 3rd Conference on Robot Learning}}.
\newblock


\bibitem[Alimohammadi et~al\mbox{.}(2024)]%
        {alimohammadi2024smite}
\bibfield{author}{\bibinfo{person}{Amirhossein Alimohammadi}, \bibinfo{person}{Sauradip Nag}, \bibinfo{person}{Saeid~Asgari Taghanaki}, \bibinfo{person}{Andrea Tagliasacchi}, \bibinfo{person}{Ghassan Hamarneh}, {and} \bibinfo{person}{Ali~Mahdavi Amiri}.} \bibinfo{year}{2024}\natexlab{}.
\newblock \showarticletitle{SMITE: Segment Me In TimE}.
\newblock \bibinfo{journal}{\emph{arXiv preprint arXiv:2410.18538}} (\bibinfo{year}{2024}).
\newblock


\bibitem[Balaji et~al\mbox{.}(2022)]%
        {balaji2022ediffi}
\bibfield{author}{\bibinfo{person}{Yogesh Balaji}, \bibinfo{person}{Seungjun Nah}, \bibinfo{person}{Xun Huang}, \bibinfo{person}{Arash Vahdat}, \bibinfo{person}{Jiaming Song}, \bibinfo{person}{Karsten Kreis}, \bibinfo{person}{Miika Aittala}, \bibinfo{person}{Timo Aila}, \bibinfo{person}{Samuli Laine}, \bibinfo{person}{Bryan Catanzaro}, {et~al\mbox{.}}} \bibinfo{year}{2022}\natexlab{}.
\newblock \showarticletitle{ediffi: Text-to-image diffusion models with an ensemble of expert denoisers}.
\newblock \bibinfo{journal}{\emph{arXiv preprint arXiv:2211.01324}} (\bibinfo{year}{2022}).
\newblock


\bibitem[Blattmann et~al\mbox{.}(2023)]%
        {blattmann2023stable}
\bibfield{author}{\bibinfo{person}{Andreas Blattmann}, \bibinfo{person}{Tim Dockhorn}, \bibinfo{person}{Sumith Kulal}, \bibinfo{person}{Daniel Mendelevitch}, \bibinfo{person}{Maciej Kilian}, \bibinfo{person}{Dominik Lorenz}, \bibinfo{person}{Yam Levi}, \bibinfo{person}{Zion English}, \bibinfo{person}{Vikram Voleti}, \bibinfo{person}{Adam Letts}, {et~al\mbox{.}}} \bibinfo{year}{2023}\natexlab{}.
\newblock \showarticletitle{Stable video diffusion: Scaling latent video diffusion models to large datasets}.
\newblock \bibinfo{journal}{\emph{arXiv preprint arXiv:2311.15127}} (\bibinfo{year}{2023}).
\newblock


\bibitem[Chang et~al\mbox{.}(2015)]%
        {chang2015shapenet}
\bibfield{author}{\bibinfo{person}{Angel~X Chang}, \bibinfo{person}{Thomas Funkhouser}, \bibinfo{person}{Leonidas Guibas}, \bibinfo{person}{Pat Hanrahan}, \bibinfo{person}{Qixing Huang}, \bibinfo{person}{Zimo Li}, \bibinfo{person}{Silvio Savarese}, \bibinfo{person}{Manolis Savva}, \bibinfo{person}{Shuran Song}, \bibinfo{person}{Hao Su}, \bibinfo{person}{Jianxiong Xiao}, \bibinfo{person}{Li Yi}, {and} \bibinfo{person}{Fisher Yu}.} \bibinfo{year}{2015}\natexlab{}.
\newblock \showarticletitle{{ShapeNet}: An information-rich {3D} model repository}.
\newblock \bibinfo{journal}{\emph{arXiv preprint arXiv:1512.03012}} (\bibinfo{year}{2015}).
\newblock


\bibitem[Collins et~al\mbox{.}(2022)]%
        {collins2022abo}
\bibfield{author}{\bibinfo{person}{Jasmine Collins}, \bibinfo{person}{Shubham Goel}, \bibinfo{person}{Kenan Deng}, \bibinfo{person}{Achleshwar Luthra}, \bibinfo{person}{Leon Xu}, \bibinfo{person}{Erhan Gundogdu}, \bibinfo{person}{Xi Zhang}, \bibinfo{person}{Tomas F~Yago Vicente}, \bibinfo{person}{Thomas Dideriksen}, \bibinfo{person}{Himanshu Arora}, \bibinfo{person}{Matthieu Guillaumin}, {and} \bibinfo{person}{Jitendra Malik}.} \bibinfo{year}{2022}\natexlab{}.
\newblock \showarticletitle{{ABO:} Dataset and benchmarks for real-world {3D} object understanding}. In \bibinfo{booktitle}{\emph{CVPR}}. \bibinfo{pages}{21126--21136}.
\newblock


\bibitem[Cong et~al\mbox{.}(2024)]%
        {FLATTEN}
\bibfield{author}{\bibinfo{person}{Yuren Cong}, \bibinfo{person}{Mengmeng Xu}, \bibinfo{person}{Christian Simon}, \bibinfo{person}{Shoufa Chen}, \bibinfo{person}{Jiawei Ren}, \bibinfo{person}{Yanping Xie}, \bibinfo{person}{Juan-Manuel Perez-Rua}, \bibinfo{person}{Bodo Rosenhahn}, \bibinfo{person}{Tao Xiang}, {and} \bibinfo{person}{Sen He}.} \bibinfo{year}{2024}\natexlab{}.
\newblock \showarticletitle{FLATTEN: optical FLow-guided ATTENtion for consistent text-to-video editing}.
\newblock  (\bibinfo{year}{2024}).
\newblock
\showeprint[arxiv]{2310.05922}~[cs.CV]
\urldef\tempurl%
\url{https://arxiv.org/abs/2310.05922}
\showURL{%
\tempurl}


\bibitem[Deitke et~al\mbox{.}(2024)]%
        {deitke2024objaverse}
\bibfield{author}{\bibinfo{person}{Matt Deitke}, \bibinfo{person}{Ruoshi Liu}, \bibinfo{person}{Matthew Wallingford}, \bibinfo{person}{Huong Ngo}, \bibinfo{person}{Oscar Michel}, \bibinfo{person}{Aditya Kusupati}, \bibinfo{person}{Alan Fan}, \bibinfo{person}{Christian Laforte}, \bibinfo{person}{Vikram Voleti}, \bibinfo{person}{Samir~Yitzhak Gadre}, {et~al\mbox{.}}} \bibinfo{year}{2024}\natexlab{}.
\newblock \showarticletitle{Objaverse-xl: A universe of 10m+ 3d objects}.
\newblock \bibinfo{journal}{\emph{Advances in Neural Information Processing Systems}}  \bibinfo{volume}{36} (\bibinfo{year}{2024}).
\newblock


\bibitem[Deitke et~al\mbox{.}(2023a)]%
        {deitke2023objaversexluniverse10m3d}
\bibfield{author}{\bibinfo{person}{Matt Deitke}, \bibinfo{person}{Ruoshi Liu}, \bibinfo{person}{Matthew Wallingford}, \bibinfo{person}{Huong Ngo}, \bibinfo{person}{Oscar Michel}, \bibinfo{person}{Aditya Kusupati}, \bibinfo{person}{Alan Fan}, \bibinfo{person}{Christian Laforte}, \bibinfo{person}{Vikram Voleti}, \bibinfo{person}{Samir~Yitzhak Gadre}, \bibinfo{person}{Eli VanderBilt}, \bibinfo{person}{Aniruddha Kembhavi}, \bibinfo{person}{Carl Vondrick}, \bibinfo{person}{Georgia Gkioxari}, \bibinfo{person}{Kiana Ehsani}, \bibinfo{person}{Ludwig Schmidt}, {and} \bibinfo{person}{Ali Farhadi}.} \bibinfo{year}{2023}\natexlab{a}.
\newblock \bibinfo{title}{Objaverse-XL: A Universe of 10M+ 3D Objects}.
\newblock
\showeprint[arxiv]{2307.05663}~[cs.CV]
\urldef\tempurl%
\url{https://arxiv.org/abs/2307.05663}
\showURL{%
\tempurl}


\bibitem[Deitke et~al\mbox{.}(2023b)]%
        {deitke2023objaverse}
\bibfield{author}{\bibinfo{person}{Matt Deitke}, \bibinfo{person}{Dustin Schwenk}, \bibinfo{person}{Jordi Salvador}, \bibinfo{person}{Luca Weihs}, \bibinfo{person}{Oscar Michel}, \bibinfo{person}{Eli VanderBilt}, \bibinfo{person}{Ludwig Schmidt}, \bibinfo{person}{Kiana Ehsani}, \bibinfo{person}{Aniruddha Kembhavi}, {and} \bibinfo{person}{Ali Farhadi}.} \bibinfo{year}{2023}\natexlab{b}.
\newblock \showarticletitle{Objaverse: A universe of annotated 3d objects}. In \bibinfo{booktitle}{\emph{Proceedings of the IEEE/CVF Conference on Computer Vision and Pattern Recognition}}. \bibinfo{pages}{13142--13153}.
\newblock


\bibitem[Fu et~al\mbox{.}(2021)]%
        {fu20213d}
\bibfield{author}{\bibinfo{person}{Huan Fu}, \bibinfo{person}{Rongfei Jia}, \bibinfo{person}{Lin Gao}, \bibinfo{person}{Mingming Gong}, \bibinfo{person}{Binqiang Zhao}, \bibinfo{person}{Steve Maybank}, {and} \bibinfo{person}{Dacheng Tao}.} \bibinfo{year}{2021}\natexlab{}.
\newblock \showarticletitle{3d-future: 3d furniture shape with texture}.
\newblock \bibinfo{journal}{\emph{International Journal of Computer Vision}}  \bibinfo{volume}{129} (\bibinfo{year}{2021}), \bibinfo{pages}{3313--3337}.
\newblock


\bibitem[Ge et~al\mbox{.}(2023)]%
        {ge2023preserve}
\bibfield{author}{\bibinfo{person}{Songwei Ge}, \bibinfo{person}{Seungjun Nah}, \bibinfo{person}{Guilin Liu}, \bibinfo{person}{Tyler Poon}, \bibinfo{person}{Andrew Tao}, \bibinfo{person}{Bryan Catanzaro}, \bibinfo{person}{David Jacobs}, \bibinfo{person}{Jia-Bin Huang}, \bibinfo{person}{Ming-Yu Liu}, {and} \bibinfo{person}{Yogesh Balaji}.} \bibinfo{year}{2023}\natexlab{}.
\newblock \showarticletitle{Preserve your own correlation: A noise prior for video diffusion models}.
\newblock \bibinfo{journal}{\emph{arXiv preprint arXiv:2305.10474}} (\bibinfo{year}{2023}).
\newblock


\bibitem[Guo et~al\mbox{.}(2023)]%
        {guo2023animatediff}
\bibfield{author}{\bibinfo{person}{Yuwei Guo}, \bibinfo{person}{Ceyuan Yang}, \bibinfo{person}{Anyi Rao}, \bibinfo{person}{Zhengyang Liang}, \bibinfo{person}{Yaohui Wang}, \bibinfo{person}{Yu Qiao}, \bibinfo{person}{Maneesh Agrawala}, \bibinfo{person}{Dahua Lin}, {and} \bibinfo{person}{Bo Dai}.} \bibinfo{year}{2023}\natexlab{}.
\newblock \showarticletitle{Animatediff: Animate your personalized text-to-image diffusion models without specific tuning}.
\newblock \bibinfo{journal}{\emph{arXiv preprint arXiv:2307.04725}} (\bibinfo{year}{2023}).
\newblock


\bibitem[Ho et~al\mbox{.}(2020)]%
        {ho2020denoising}
\bibfield{author}{\bibinfo{person}{Jonathan Ho}, \bibinfo{person}{Ajay Jain}, {and} \bibinfo{person}{Pieter Abbeel}.} \bibinfo{year}{2020}\natexlab{}.
\newblock \showarticletitle{Denoising diffusion probabilistic models}.
\newblock \bibinfo{journal}{\emph{Advances in neural information processing systems}}  \bibinfo{volume}{33} (\bibinfo{year}{2020}), \bibinfo{pages}{6840--6851}.
\newblock


\bibitem[Hong et~al\mbox{.}(2022)]%
        {hong2022cogvideo}
\bibfield{author}{\bibinfo{person}{Wenyi Hong}, \bibinfo{person}{Ming Ding}, \bibinfo{person}{Wendi Zheng}, \bibinfo{person}{Xinghan Liu}, {and} \bibinfo{person}{Jie Tang}.} \bibinfo{year}{2022}\natexlab{}.
\newblock \showarticletitle{Cogvideo: Large-scale pretraining for text-to-video generation via transformers}.
\newblock \bibinfo{journal}{\emph{arXiv preprint arXiv:2205.15868}} (\bibinfo{year}{2022}).
\newblock


\bibitem[Hu et~al\mbox{.}(2017)]%
        {hu2017learning}
\bibfield{author}{\bibinfo{person}{Ruizhen Hu}, \bibinfo{person}{Wenchao Li}, \bibinfo{person}{Oliver Van~Kaick}, \bibinfo{person}{Ariel Shamir}, \bibinfo{person}{Hao Zhang}, {and} \bibinfo{person}{Hui Huang}.} \bibinfo{year}{2017}\natexlab{}.
\newblock \showarticletitle{Learning to predict part mobility from a single static snapshot}.
\newblock \bibinfo{journal}{\emph{ACM Transactions On Graphics (TOG)}} \bibinfo{volume}{36}, \bibinfo{number}{6} (\bibinfo{year}{2017}), \bibinfo{pages}{1--13}.
\newblock


\bibitem[Iliash et~al\mbox{.}(2024)]%
        {iliash2024s2o}
\bibfield{author}{\bibinfo{person}{Denys Iliash}, \bibinfo{person}{Hanxiao Jiang}, \bibinfo{person}{Yiming Zhang}, \bibinfo{person}{Manolis Savva}, {and} \bibinfo{person}{Angel~X Chang}.} \bibinfo{year}{2024}\natexlab{}.
\newblock \showarticletitle{S2O: Static to openable enhancement for articulated 3D objects}.
\newblock \bibinfo{journal}{\emph{arXiv preprint arXiv:2409.18896}} (\bibinfo{year}{2024}).
\newblock


\bibitem[Jain et~al\mbox{.}(2021)]%
        {jain2021screwnet}
\bibfield{author}{\bibinfo{person}{Ajinkya Jain}, \bibinfo{person}{Rudolf Lioutikov}, \bibinfo{person}{Caleb Chuck}, {and} \bibinfo{person}{Scott Niekum}.} \bibinfo{year}{2021}\natexlab{}.
\newblock \showarticletitle{Screwnet: Category-independent articulation model estimation from depth images using screw theory}. In \bibinfo{booktitle}{\emph{2021 IEEE International Conference on Robotics and Automation (ICRA)}}. IEEE, \bibinfo{pages}{13670--13677}.
\newblock


\bibitem[Jeong et~al\mbox{.}(2023)]%
        {jeong2023vmc}
\bibfield{author}{\bibinfo{person}{Hyeonho Jeong}, \bibinfo{person}{Geon~Yeong Park}, {and} \bibinfo{person}{Jong~Chul Ye}.} \bibinfo{year}{2023}\natexlab{}.
\newblock \showarticletitle{VMC: Video Motion Customization using Temporal Attention Adaption for Text-to-Video Diffusion Models}.
\newblock \bibinfo{journal}{\emph{arXiv preprint arXiv:2312.00845}} (\bibinfo{year}{2023}).
\newblock


\bibitem[Jiang et~al\mbox{.}(2022b)]%
        {jiang2022opd}
\bibfield{author}{\bibinfo{person}{Hanxiao Jiang}, \bibinfo{person}{Yongsen Mao}, \bibinfo{person}{Manolis Savva}, {and} \bibinfo{person}{Angel~X Chang}.} \bibinfo{year}{2022}\natexlab{b}.
\newblock \showarticletitle{OPD: Single-view 3D openable part detection}. In \bibinfo{booktitle}{\emph{European Conference on Computer Vision}}. Springer, \bibinfo{pages}{410--426}.
\newblock


\bibitem[Jiang et~al\mbox{.}(2022a)]%
        {jiang2022ditto}
\bibfield{author}{\bibinfo{person}{Zhenyu Jiang}, \bibinfo{person}{Cheng-Chun Hsu}, {and} \bibinfo{person}{Yuke Zhu}.} \bibinfo{year}{2022}\natexlab{a}.
\newblock \showarticletitle{Ditto: Building digital twins of articulated objects from interaction}. In \bibinfo{booktitle}{\emph{Proceedings of the IEEE/CVF Conference on Computer Vision and Pattern Recognition}}. \bibinfo{pages}{5616--5626}.
\newblock


\bibitem[Keselman and Hebert(2022)]%
        {keselman2022approximate}
\bibfield{author}{\bibinfo{person}{Leonid Keselman} {and} \bibinfo{person}{Martial Hebert}.} \bibinfo{year}{2022}\natexlab{}.
\newblock \showarticletitle{Approximate differentiable rendering with algebraic surfaces}. In \bibinfo{booktitle}{\emph{European Conference on Computer Vision}}. Springer, \bibinfo{pages}{596--614}.
\newblock


\bibitem[Keselman and Hebert(2023)]%
        {keselman2023flexible}
\bibfield{author}{\bibinfo{person}{Leonid Keselman} {and} \bibinfo{person}{Martial Hebert}.} \bibinfo{year}{2023}\natexlab{}.
\newblock \showarticletitle{Flexible techniques for differentiable rendering with 3d gaussians}.
\newblock \bibinfo{journal}{\emph{arXiv preprint arXiv:2308.14737}} (\bibinfo{year}{2023}).
\newblock


\bibitem[Khanna et~al\mbox{.}(2024)]%
        {khanna2024habitat}
\bibfield{author}{\bibinfo{person}{Mukul Khanna}, \bibinfo{person}{Yongsen Mao}, \bibinfo{person}{Hanxiao Jiang}, \bibinfo{person}{Sanjay Haresh}, \bibinfo{person}{Brennan Shacklett}, \bibinfo{person}{Dhruv Batra}, \bibinfo{person}{Alexander Clegg}, \bibinfo{person}{Eric Undersander}, \bibinfo{person}{Angel~X Chang}, {and} \bibinfo{person}{Manolis Savva}.} \bibinfo{year}{2024}\natexlab{}.
\newblock \showarticletitle{Habitat synthetic scenes dataset (hssd-200): An analysis of 3d scene scale and realism tradeoffs for objectgoal navigation}. In \bibinfo{booktitle}{\emph{Proceedings of the IEEE/CVF Conference on Computer Vision and Pattern Recognition}}. \bibinfo{pages}{16384--16393}.
\newblock


\bibitem[Kim and Sung(2024)]%
        {kim2024partstad}
\bibfield{author}{\bibinfo{person}{Hyunjin Kim} {and} \bibinfo{person}{Minhyuk Sung}.} \bibinfo{year}{2024}\natexlab{}.
\newblock \showarticletitle{Partstad: 2d-to-3d part segmentation task adaptation}. In \bibinfo{booktitle}{\emph{European Conference on Computer Vision}}. Springer, \bibinfo{pages}{422--439}.
\newblock


\bibitem[Kingma and Ba(2014)]%
        {kingma2014adam}
\bibfield{author}{\bibinfo{person}{Diederik~P Kingma} {and} \bibinfo{person}{Jimmy Ba}.} \bibinfo{year}{2014}\natexlab{}.
\newblock \showarticletitle{Adam: A method for stochastic optimization}.
\newblock \bibinfo{journal}{\emph{arXiv preprint arXiv:1412.6980}} (\bibinfo{year}{2014}).
\newblock


\bibitem[Kingma and Welling(2013)]%
        {vae}
\bibfield{author}{\bibinfo{person}{Diederik~P Kingma} {and} \bibinfo{person}{Max Welling}.} \bibinfo{year}{2013}\natexlab{}.
\newblock \showarticletitle{Auto-encoding variational bayes}.
\newblock \bibinfo{journal}{\emph{arXiv preprint arXiv:1312.6114}} (\bibinfo{year}{2013}).
\newblock


\bibitem[Le~Moing et~al\mbox{.}(2021)]%
        {le2021ccvs}
\bibfield{author}{\bibinfo{person}{Guillaume Le~Moing}, \bibinfo{person}{Jean Ponce}, {and} \bibinfo{person}{Cordelia Schmid}.} \bibinfo{year}{2021}\natexlab{}.
\newblock \showarticletitle{CCVS: context-aware controllable video synthesis}.
\newblock \bibinfo{journal}{\emph{Advances in Neural Information Processing Systems}}  \bibinfo{volume}{34} (\bibinfo{year}{2021}), \bibinfo{pages}{14042--14055}.
\newblock


\bibitem[Lei et~al\mbox{.}(2023)]%
        {lei2023nap}
\bibfield{author}{\bibinfo{person}{Jiahui Lei}, \bibinfo{person}{Congyue Deng}, \bibinfo{person}{William~B Shen}, \bibinfo{person}{Leonidas~J Guibas}, {and} \bibinfo{person}{Kostas Daniilidis}.} \bibinfo{year}{2023}\natexlab{}.
\newblock \showarticletitle{Nap: Neural 3d articulated object prior}.
\newblock \bibinfo{journal}{\emph{Advances in Neural Information Processing Systems}}  \bibinfo{volume}{36} (\bibinfo{year}{2023}), \bibinfo{pages}{31878--31894}.
\newblock


\bibitem[Li et~al\mbox{.}(2024)]%
        {li2024vivid}
\bibfield{author}{\bibinfo{person}{Bing Li}, \bibinfo{person}{Cheng Zheng}, \bibinfo{person}{Wenxuan Zhu}, \bibinfo{person}{Jinjie Mai}, \bibinfo{person}{Biao Zhang}, \bibinfo{person}{Peter Wonka}, {and} \bibinfo{person}{Bernard Ghanem}.} \bibinfo{year}{2024}\natexlab{}.
\newblock \showarticletitle{Vivid-zoo: Multi-view video generation with diffusion model}.
\newblock \bibinfo{journal}{\emph{Advances in Neural Information Processing Systems}}  \bibinfo{volume}{37} (\bibinfo{year}{2024}), \bibinfo{pages}{62189--62222}.
\newblock


\bibitem[Li et~al\mbox{.}(2025)]%
        {li2025puppet}
\bibfield{author}{\bibinfo{person}{Ruining Li}, \bibinfo{person}{Chuanxia Zheng}, \bibinfo{person}{Christian Rupprecht}, {and} \bibinfo{person}{Andrea Vedaldi}.} \bibinfo{year}{2025}\natexlab{}.
\newblock \showarticletitle{Puppet-master: Scaling interactive video generation as a motion prior for part-level dynamics}. In \bibinfo{booktitle}{\emph{ICCV}}.
\newblock


\bibitem[Liu et~al\mbox{.}(2024a)]%
        {liu2024singapo}
\bibfield{author}{\bibinfo{person}{Jiayi Liu}, \bibinfo{person}{Denys Iliash}, \bibinfo{person}{Angel~X Chang}, \bibinfo{person}{Manolis Savva}, {and} \bibinfo{person}{Ali Mahdavi-Amiri}.} \bibinfo{year}{2024}\natexlab{a}.
\newblock \showarticletitle{SINGAPO: Single Image Controlled Generation of Articulated Parts in Object}.
\newblock \bibinfo{journal}{\emph{arXiv preprint arXiv:2410.16499}} (\bibinfo{year}{2024}).
\newblock


\bibitem[Liu et~al\mbox{.}(2023a)]%
        {liu2023paris}
\bibfield{author}{\bibinfo{person}{Jiayi Liu}, \bibinfo{person}{Ali Mahdavi-Amiri}, {and} \bibinfo{person}{Manolis Savva}.} \bibinfo{year}{2023}\natexlab{a}.
\newblock \showarticletitle{Paris: Part-level reconstruction and motion analysis for articulated objects}. In \bibinfo{booktitle}{\emph{Proceedings of the IEEE/CVF International Conference on Computer Vision}}. \bibinfo{pages}{352--363}.
\newblock


\bibitem[Liu et~al\mbox{.}(2024b)]%
        {liu2024cage}
\bibfield{author}{\bibinfo{person}{Jiayi Liu}, \bibinfo{person}{Hou In~Ivan Tam}, \bibinfo{person}{Ali Mahdavi-Amiri}, {and} \bibinfo{person}{Manolis Savva}.} \bibinfo{year}{2024}\natexlab{b}.
\newblock \showarticletitle{CAGE: Controllable Articulation GEneration}. In \bibinfo{booktitle}{\emph{Proceedings of the IEEE/CVF Conference on Computer Vision and Pattern Recognition}}. \bibinfo{pages}{17880--17889}.
\newblock


\bibitem[Liu et~al\mbox{.}(2023b)]%
        {liu2023partslip}
\bibfield{author}{\bibinfo{person}{Minghua Liu}, \bibinfo{person}{Yinhao Zhu}, \bibinfo{person}{Hong Cai}, \bibinfo{person}{Shizhong Han}, \bibinfo{person}{Zhan Ling}, \bibinfo{person}{Fatih Porikli}, {and} \bibinfo{person}{Hao Su}.} \bibinfo{year}{2023}\natexlab{b}.
\newblock \showarticletitle{Partslip: Low-shot part segmentation for 3d point clouds via pretrained image-language models}. In \bibinfo{booktitle}{\emph{Proceedings of the IEEE/CVF Conference on Computer Vision and Pattern Recognition}}. \bibinfo{pages}{21736--21746}.
\newblock


\bibitem[Luo et~al\mbox{.}(2023)]%
        {luo2023videofusion}
\bibfield{author}{\bibinfo{person}{Zhengxiong Luo}, \bibinfo{person}{Dayou Chen}, \bibinfo{person}{Yingya Zhang}, \bibinfo{person}{Yan Huang}, \bibinfo{person}{Liang Wang}, \bibinfo{person}{Yujun Shen}, \bibinfo{person}{Deli Zhao}, \bibinfo{person}{Jingren Zhou}, {and} \bibinfo{person}{Tieniu Tan}.} \bibinfo{year}{2023}\natexlab{}.
\newblock \showarticletitle{VideoFusion: Decomposed Diffusion Models for High-Quality Video Generation}. In \bibinfo{booktitle}{\emph{Proceedings of the IEEE/CVF Conference on Computer Vision and Pattern Recognition}}. \bibinfo{pages}{10209--10218}.
\newblock


\bibitem[Mandi et~al\mbox{.}(2024)]%
        {mandi2024real2code}
\bibfield{author}{\bibinfo{person}{Zhao Mandi}, \bibinfo{person}{Yijia Weng}, \bibinfo{person}{Dominik Bauer}, {and} \bibinfo{person}{Shuran Song}.} \bibinfo{year}{2024}\natexlab{}.
\newblock \showarticletitle{Real2code: Reconstruct articulated objects via code generation}.
\newblock \bibinfo{journal}{\emph{arXiv preprint arXiv:2406.08474}} (\bibinfo{year}{2024}).
\newblock


\bibitem[Materzynska et~al\mbox{.}(2023)]%
        {materzynska2023customizing}
\bibfield{author}{\bibinfo{person}{Joanna Materzynska}, \bibinfo{person}{Josef Sivic}, \bibinfo{person}{Eli Shechtman}, \bibinfo{person}{Antonio Torralba}, \bibinfo{person}{Richard Zhang}, {and} \bibinfo{person}{Bryan Russell}.} \bibinfo{year}{2023}\natexlab{}.
\newblock \showarticletitle{Customizing motion in text-to-video diffusion models}.
\newblock \bibinfo{journal}{\emph{arXiv preprint arXiv:2312.04966}} (\bibinfo{year}{2023}).
\newblock


\bibitem[Miller et~al\mbox{.}(2020)]%
        {CLIP1}
\bibfield{author}{\bibinfo{person}{Alexander~H. Miller}, \bibinfo{person}{Will Feng}, \bibinfo{person}{Dhruva Tirumala}, \bibinfo{person}{Adam Fisch}, \bibinfo{person}{Augustus Odena}, \bibinfo{person}{Vivek Ramavajjala}, \bibinfo{person}{Joel~Z. Leibo}, \bibinfo{person}{Kelvin~Guu andJesse Engel}, \bibinfo{person}{Jack Clark}, \bibinfo{person}{Maruan~H. Ali}, \bibinfo{person}{Nazneen Rajani}, \bibinfo{person}{Iain~J. Dunning}, \bibinfo{person}{Jacob Andreas}, \bibinfo{person}{Chris Dyer}, \bibinfo{person}{Dario Amodei}, \bibinfo{person}{Jakob Uszkoreit}, \bibinfo{person}{Douwe Pieksma}, \bibinfo{person}{Tom Brown}, {and} \bibinfo{person}{Ilya Sutskever}.} \bibinfo{year}{2020}\natexlab{}.
\newblock \showarticletitle{CLIP: Learning to Solve Visual Tasks by Unsupervised Learning of Language Representations}. In \bibinfo{booktitle}{\emph{International Conference on Machine Learning}}.
\newblock


\bibitem[Mo et~al\mbox{.}(2021)]%
        {mo2021where2act}
\bibfield{author}{\bibinfo{person}{Kaichun Mo}, \bibinfo{person}{Leonidas~J Guibas}, \bibinfo{person}{Mustafa Mukadam}, \bibinfo{person}{Abhinav Gupta}, {and} \bibinfo{person}{Shubham Tulsiani}.} \bibinfo{year}{2021}\natexlab{}.
\newblock \showarticletitle{Where2act: From pixels to actions for articulated 3d objects}. In \bibinfo{booktitle}{\emph{Proceedings of the IEEE/CVF International Conference on Computer Vision}}. \bibinfo{pages}{6813--6823}.
\newblock


\bibitem[Mo et~al\mbox{.}(2019)]%
        {mo2019partnet}
\bibfield{author}{\bibinfo{person}{Kaichun Mo}, \bibinfo{person}{Shilin Zhu}, \bibinfo{person}{Angel~X. Chang}, \bibinfo{person}{Li Yi}, \bibinfo{person}{Subarna Tripathi}, \bibinfo{person}{Leonidas~J. Guibas}, {and} \bibinfo{person}{Hao Su}.} \bibinfo{year}{2019}\natexlab{}.
\newblock \showarticletitle{{PartNet}: A Large-Scale Benchmark for Fine-Grained and Hierarchical Part-Level {3D} Object Understanding}. In \bibinfo{booktitle}{\emph{The IEEE Conference on Computer Vision and Pattern Recognition (CVPR)}}.
\newblock


\bibitem[Molad et~al\mbox{.}(2023)]%
        {dreamix}
\bibfield{author}{\bibinfo{person}{Eyal Molad}, \bibinfo{person}{Eliahu Horwitz}, \bibinfo{person}{Dani Valevski}, \bibinfo{person}{Alex~Rav Acha}, \bibinfo{person}{Yossi Matias}, \bibinfo{person}{Yael Pritch}, \bibinfo{person}{Yaniv Leviathan}, {and} \bibinfo{person}{Yedid Hoshen}.} \bibinfo{year}{2023}\natexlab{}.
\newblock \showarticletitle{Dreamix: Video Diffusion Models are General Video Editors}.
\newblock \bibinfo{journal}{\emph{arXiv preprint arXiv:2302.01329}} (\bibinfo{year}{2023}).
\newblock


\bibitem[Nag et~al\mbox{.}(2025)]%
        {nag20252}
\bibfield{author}{\bibinfo{person}{Sauradip Nag}, \bibinfo{person}{Daniel Cohen-Or}, \bibinfo{person}{Hao Zhang}, {and} \bibinfo{person}{Ali Mahdavi-Amiri}.} \bibinfo{year}{2025}\natexlab{}.
\newblock \showarticletitle{In-2-4d: Inbetweening from two single-view images to 4d generation}.
\newblock \bibinfo{journal}{\emph{arXiv preprint arXiv:2504.08366}} (\bibinfo{year}{2025}).
\newblock


\bibitem[Nag et~al\mbox{.}(2023)]%
        {nag2023difftad}
\bibfield{author}{\bibinfo{person}{Sauradip Nag}, \bibinfo{person}{Xiatian Zhu}, \bibinfo{person}{Jiankang Deng}, \bibinfo{person}{Yi-Zhe Song}, {and} \bibinfo{person}{Tao Xiang}.} \bibinfo{year}{2023}\natexlab{}.
\newblock \showarticletitle{Difftad: Temporal action detection with proposal denoising diffusion}. In \bibinfo{booktitle}{\emph{Proceedings of the IEEE/CVF International Conference on Computer Vision}}. \bibinfo{pages}{10362--10374}.
\newblock


\bibitem[Paszke et~al\mbox{.}(2019)]%
        {paszke2019pytorch}
\bibfield{author}{\bibinfo{person}{Adam Paszke}, \bibinfo{person}{Sam Gross}, \bibinfo{person}{Francisco Massa}, \bibinfo{person}{Adam Lerer}, \bibinfo{person}{James Bradbury}, \bibinfo{person}{Gregory Chanan}, \bibinfo{person}{Trevor Killeen}, \bibinfo{person}{Zeming Lin}, \bibinfo{person}{Natalia Gimelshein}, \bibinfo{person}{Luca Antiga}, {et~al\mbox{.}}} \bibinfo{year}{2019}\natexlab{}.
\newblock \showarticletitle{Pytorch: An imperative style, high-performance deep learning library}.
\newblock \bibinfo{journal}{\emph{Advances in neural information processing systems}}  \bibinfo{volume}{32} (\bibinfo{year}{2019}).
\newblock


\bibitem[Perla et~al\mbox{.}(2025)]%
        {perla2025asia}
\bibfield{author}{\bibinfo{person}{Sai Raj~Kishore Perla}, \bibinfo{person}{Aditya Vora}, \bibinfo{person}{Sauradip Nag}, \bibinfo{person}{Ali Mahdavi-Amiri}, {and} \bibinfo{person}{Hao Zhang}.} \bibinfo{year}{2025}\natexlab{}.
\newblock \showarticletitle{ASIA: Adaptive 3D Segmentation using Few Image Annotations}. In \bibinfo{booktitle}{\emph{Proceedings of the SIGGRAPH Asia 2025 Conference Papers}}. \bibinfo{pages}{1--12}.
\newblock


\bibitem[Podell et~al\mbox{.}(2023)]%
        {podell2023sdxl}
\bibfield{author}{\bibinfo{person}{Dustin Podell}, \bibinfo{person}{Zion English}, \bibinfo{person}{Kyle Lacey}, \bibinfo{person}{Andreas Blattmann}, \bibinfo{person}{Tim Dockhorn}, \bibinfo{person}{Jonas M{\"u}ller}, \bibinfo{person}{Joe Penna}, {and} \bibinfo{person}{Robin Rombach}.} \bibinfo{year}{2023}\natexlab{}.
\newblock \showarticletitle{Sdxl: Improving latent diffusion models for high-resolution image synthesis}.
\newblock \bibinfo{journal}{\emph{arXiv preprint arXiv:2307.01952}} (\bibinfo{year}{2023}).
\newblock


\bibitem[Poole et~al\mbox{.}(2022)]%
        {poole2022dreamfusion}
\bibfield{author}{\bibinfo{person}{Ben Poole}, \bibinfo{person}{Ajay Jain}, \bibinfo{person}{Jonathan~T. Barron}, {and} \bibinfo{person}{Ben Mildenhall}.} \bibinfo{year}{2022}\natexlab{}.
\newblock \showarticletitle{DreamFusion: Text-to-3D using 2D Diffusion}.
\newblock \bibinfo{journal}{\emph{arXiv}} (\bibinfo{year}{2022}).
\newblock


\bibitem[Qi et~al\mbox{.}(2023)]%
        {qi2023fatezero}
\bibfield{author}{\bibinfo{person}{Chenyang Qi}, \bibinfo{person}{Xiaodong Cun}, \bibinfo{person}{Yong Zhang}, \bibinfo{person}{Chenyang Lei}, \bibinfo{person}{Xintao Wang}, \bibinfo{person}{Ying Shan}, {and} \bibinfo{person}{Qifeng Chen}.} \bibinfo{year}{2023}\natexlab{}.
\newblock \showarticletitle{Fatezero: Fusing attentions for zero-shot text-based video editing}. In \bibinfo{booktitle}{\emph{Proceedings of the IEEE/CVF International Conference on Computer Vision}}. \bibinfo{pages}{15932--15942}.
\newblock


\bibitem[Qiu et~al\mbox{.}(2025)]%
        {qiu2025articulate}
\bibfield{author}{\bibinfo{person}{Xiaowen Qiu}, \bibinfo{person}{Jincheng Yang}, \bibinfo{person}{Yian Wang}, \bibinfo{person}{Zhehuan Chen}, \bibinfo{person}{Yufei Wang}, \bibinfo{person}{Tsun-Hsuan Wang}, \bibinfo{person}{Zhou Xian}, {and} \bibinfo{person}{Chuang Gan}.} \bibinfo{year}{2025}\natexlab{}.
\newblock \showarticletitle{Articulate AnyMesh: Open-Vocabulary 3D Articulated Objects Modeling}.
\newblock \bibinfo{journal}{\emph{arXiv preprint arXiv:2502.02590}} (\bibinfo{year}{2025}).
\newblock


\bibitem[Ramesh et~al\mbox{.}(2021)]%
        {ramesh2021zero}
\bibfield{author}{\bibinfo{person}{Aditya Ramesh}, \bibinfo{person}{Mikhail Pavlov}, \bibinfo{person}{Gabriel Goh}, \bibinfo{person}{Scott Gray}, \bibinfo{person}{Chelsea Voss}, \bibinfo{person}{Alec Radford}, \bibinfo{person}{Mark Chen}, {and} \bibinfo{person}{Ilya Sutskever}.} \bibinfo{year}{2021}\natexlab{}.
\newblock \showarticletitle{Zero-shot text-to-image generation}. In \bibinfo{booktitle}{\emph{International Conference on Machine Learning}}. PMLR, \bibinfo{pages}{8821--8831}.
\newblock


\bibitem[Ravi et~al\mbox{.}(2024)]%
        {ravi2024sam2segmentimages}
\bibfield{author}{\bibinfo{person}{Nikhila Ravi}, \bibinfo{person}{Valentin Gabeur}, \bibinfo{person}{Yuan-Ting Hu}, \bibinfo{person}{Ronghang Hu}, \bibinfo{person}{Chaitanya Ryali}, \bibinfo{person}{Tengyu Ma}, \bibinfo{person}{Haitham Khedr}, \bibinfo{person}{Roman Rädle}, \bibinfo{person}{Chloe Rolland}, \bibinfo{person}{Laura Gustafson}, \bibinfo{person}{Eric Mintun}, \bibinfo{person}{Junting Pan}, \bibinfo{person}{Kalyan~Vasudev Alwala}, \bibinfo{person}{Nicolas Carion}, \bibinfo{person}{Chao-Yuan Wu}, \bibinfo{person}{Ross Girshick}, \bibinfo{person}{Piotr Dollár}, {and} \bibinfo{person}{Christoph Feichtenhofer}.} \bibinfo{year}{2024}\natexlab{}.
\newblock \bibinfo{title}{SAM 2: Segment Anything in Images and Videos}.
\newblock
\showeprint[arxiv]{2408.00714}~[cs.CV]
\urldef\tempurl%
\url{https://arxiv.org/abs/2408.00714}
\showURL{%
\tempurl}


\bibitem[Ren et~al\mbox{.}(2024)]%
        {ren2024l4gm}
\bibfield{author}{\bibinfo{person}{Jiawei Ren}, \bibinfo{person}{Cheng Xie}, \bibinfo{person}{Ashkan Mirzaei}, \bibinfo{person}{Karsten Kreis}, \bibinfo{person}{Ziwei Liu}, \bibinfo{person}{Antonio Torralba}, \bibinfo{person}{Sanja Fidler}, \bibinfo{person}{Seung~Wook Kim}, \bibinfo{person}{Huan Ling}, {et~al\mbox{.}}} \bibinfo{year}{2024}\natexlab{}.
\newblock \showarticletitle{L4gm: Large 4d gaussian reconstruction model}.
\newblock \bibinfo{journal}{\emph{Advances in Neural Information Processing Systems}}  \bibinfo{volume}{37} (\bibinfo{year}{2024}), \bibinfo{pages}{56828--56858}.
\newblock


\bibitem[Rombach et~al\mbox{.}(2022)]%
        {rombach2022high}
\bibfield{author}{\bibinfo{person}{Robin Rombach}, \bibinfo{person}{Andreas Blattmann}, \bibinfo{person}{Dominik Lorenz}, \bibinfo{person}{Patrick Esser}, {and} \bibinfo{person}{Bj{\"o}rn Ommer}.} \bibinfo{year}{2022}\natexlab{}.
\newblock \showarticletitle{High-resolution image synthesis with latent diffusion models}. In \bibinfo{booktitle}{\emph{Proceedings of the IEEE/CVF conference on computer vision and pattern recognition}}. \bibinfo{pages}{10684--10695}.
\newblock


\bibitem[Ronneberger et~al\mbox{.}(2015)]%
        {ronneberger2015u}
\bibfield{author}{\bibinfo{person}{Olaf Ronneberger}, \bibinfo{person}{Philipp Fischer}, {and} \bibinfo{person}{Thomas Brox}.} \bibinfo{year}{2015}\natexlab{}.
\newblock \showarticletitle{U-net: Convolutional networks for biomedical image segmentation}. In \bibinfo{booktitle}{\emph{Medical image computing and computer-assisted intervention--MICCAI 2015: 18th international conference, Munich, Germany, October 5-9, 2015, proceedings, part III 18}}. Springer, \bibinfo{pages}{234--241}.
\newblock


\bibitem[Saharia et~al\mbox{.}(2022)]%
        {saharia2022photorealistic}
\bibfield{author}{\bibinfo{person}{Chitwan Saharia}, \bibinfo{person}{William Chan}, \bibinfo{person}{Saurabh Saxena}, \bibinfo{person}{Lala Li}, \bibinfo{person}{Jay Whang}, \bibinfo{person}{Emily~L Denton}, \bibinfo{person}{Kamyar Ghasemipour}, \bibinfo{person}{Raphael Gontijo~Lopes}, \bibinfo{person}{Burcu Karagol~Ayan}, \bibinfo{person}{Tim Salimans}, {et~al\mbox{.}}} \bibinfo{year}{2022}\natexlab{}.
\newblock \showarticletitle{Photorealistic text-to-image diffusion models with deep language understanding}.
\newblock \bibinfo{journal}{\emph{Advances in Neural Information Processing Systems}}  \bibinfo{volume}{35} (\bibinfo{year}{2022}), \bibinfo{pages}{36479--36494}.
\newblock


\bibitem[Singer et~al\mbox{.}(2022)]%
        {singer2022make}
\bibfield{author}{\bibinfo{person}{Uriel Singer}, \bibinfo{person}{Adam Polyak}, \bibinfo{person}{Thomas Hayes}, \bibinfo{person}{Xi Yin}, \bibinfo{person}{Jie An}, \bibinfo{person}{Songyang Zhang}, \bibinfo{person}{Qiyuan Hu}, \bibinfo{person}{Harry Yang}, \bibinfo{person}{Oron Ashual}, \bibinfo{person}{Oran Gafni}, {et~al\mbox{.}}} \bibinfo{year}{2022}\natexlab{}.
\newblock \showarticletitle{Make-a-video: Text-to-video generation without text-video data}.
\newblock \bibinfo{journal}{\emph{arXiv preprint arXiv:2209.14792}} (\bibinfo{year}{2022}).
\newblock


\bibitem[Song et~al\mbox{.}(2024)]%
        {song2024reacto}
\bibfield{author}{\bibinfo{person}{Chaoyue Song}, \bibinfo{person}{Jiacheng Wei}, \bibinfo{person}{Chuan~Sheng Foo}, \bibinfo{person}{Guosheng Lin}, {and} \bibinfo{person}{Fayao Liu}.} \bibinfo{year}{2024}\natexlab{}.
\newblock \showarticletitle{REACTO: Reconstructing Articulated Objects from a Single Video}. In \bibinfo{booktitle}{\emph{Proceedings of the IEEE/CVF Conference on Computer Vision and Pattern Recognition}}. \bibinfo{pages}{5384--5395}.
\newblock


\bibitem[Song et~al\mbox{.}(2020a)]%
        {song2020denoising}
\bibfield{author}{\bibinfo{person}{Jiaming Song}, \bibinfo{person}{Chenlin Meng}, {and} \bibinfo{person}{Stefano Ermon}.} \bibinfo{year}{2020}\natexlab{a}.
\newblock \showarticletitle{Denoising diffusion implicit models}.
\newblock \bibinfo{journal}{\emph{arXiv preprint arXiv:2010.02502}} (\bibinfo{year}{2020}).
\newblock


\bibitem[Song et~al\mbox{.}(2020b)]%
        {song2020score}
\bibfield{author}{\bibinfo{person}{Yang Song}, \bibinfo{person}{Jascha Sohl-Dickstein}, \bibinfo{person}{Diederik~P Kingma}, \bibinfo{person}{Abhishek Kumar}, \bibinfo{person}{Stefano Ermon}, {and} \bibinfo{person}{Ben Poole}.} \bibinfo{year}{2020}\natexlab{b}.
\newblock \showarticletitle{Score-based generative modeling through stochastic differential equations}.
\newblock \bibinfo{journal}{\emph{arXiv preprint arXiv:2011.13456}} (\bibinfo{year}{2020}).
\newblock


\bibitem[Unterthiner et~al\mbox{.}(2019)]%
        {unterthiner2019fvd}
\bibfield{author}{\bibinfo{person}{Thomas Unterthiner}, \bibinfo{person}{Sjoerd van Steenkiste}, \bibinfo{person}{Karol Kurach}, \bibinfo{person}{Rapha{\"e}l Marinier}, \bibinfo{person}{Marcin Michalski}, {and} \bibinfo{person}{Sylvain Gelly}.} \bibinfo{year}{2019}\natexlab{}.
\newblock \showarticletitle{FVD: A new metric for video generation}.
\newblock  (\bibinfo{year}{2019}).
\newblock


\bibitem[Uzolas et~al\mbox{.}(2024)]%
        {uzolas2024motiondreamer}
\bibfield{author}{\bibinfo{person}{Lukas Uzolas}, \bibinfo{person}{Elmar Eisemann}, {and} \bibinfo{person}{Petr Kellnhofer}.} \bibinfo{year}{2024}\natexlab{}.
\newblock \showarticletitle{MotionDreamer: Zero-Shot 3D Mesh Animation from Video Diffusion Models}.
\newblock \bibinfo{journal}{\emph{arXiv preprint arXiv:2405.20155}} (\bibinfo{year}{2024}).
\newblock


\bibitem[Vaswani et~al\mbox{.}(2017)]%
        {vaswani2017attention}
\bibfield{author}{\bibinfo{person}{Ashish Vaswani}, \bibinfo{person}{Noam Shazeer}, \bibinfo{person}{Niki Parmar}, \bibinfo{person}{Jakob Uszkoreit}, \bibinfo{person}{Llion Jones}, \bibinfo{person}{Aidan~N Gomez}, \bibinfo{person}{{\L}ukasz Kaiser}, {and} \bibinfo{person}{Illia Polosukhin}.} \bibinfo{year}{2017}\natexlab{}.
\newblock \showarticletitle{Attention is all you need}.
\newblock \bibinfo{journal}{\emph{Advances in neural information processing systems}}  \bibinfo{volume}{30} (\bibinfo{year}{2017}).
\newblock


\bibitem[Voleti et~al\mbox{.}(2025)]%
        {voleti2025sv3d}
\bibfield{author}{\bibinfo{person}{Vikram Voleti}, \bibinfo{person}{Chun-Han Yao}, \bibinfo{person}{Mark Boss}, \bibinfo{person}{Adam Letts}, \bibinfo{person}{David Pankratz}, \bibinfo{person}{Dmitry Tochilkin}, \bibinfo{person}{Christian Laforte}, \bibinfo{person}{Robin Rombach}, {and} \bibinfo{person}{Varun Jampani}.} \bibinfo{year}{2025}\natexlab{}.
\newblock \showarticletitle{Sv3d: Novel multi-view synthesis and 3d generation from a single image using latent video diffusion}. In \bibinfo{booktitle}{\emph{European Conference on Computer Vision}}. Springer, \bibinfo{pages}{439--457}.
\newblock


\bibitem[von Platen et~al\mbox{.}(2022)]%
        {von-platen-etal-2022-diffusers}
\bibfield{author}{\bibinfo{person}{Patrick von Platen}, \bibinfo{person}{Suraj Patil}, \bibinfo{person}{Anton Lozhkov}, \bibinfo{person}{Pedro Cuenca}, \bibinfo{person}{Nathan Lambert}, \bibinfo{person}{Kashif Rasul}, \bibinfo{person}{Mishig Davaadorj}, \bibinfo{person}{Dhruv Nair}, \bibinfo{person}{Sayak Paul}, \bibinfo{person}{William Berman}, \bibinfo{person}{Yiyi Xu}, \bibinfo{person}{Steven Liu}, {and} \bibinfo{person}{Thomas Wolf}.} \bibinfo{year}{2022}\natexlab{}.
\newblock \bibinfo{title}{Diffusers: State-of-the-art diffusion models}.
\newblock \bibinfo{howpublished}{\url{https://github.com/huggingface/diffusers}}.
\newblock


\bibitem[Wan et~al\mbox{.}(2025)]%
        {wan2025wan}
\bibfield{author}{\bibinfo{person}{Team Wan}, \bibinfo{person}{Ang Wang}, \bibinfo{person}{Baole Ai}, \bibinfo{person}{Bin Wen}, \bibinfo{person}{Chaojie Mao}, \bibinfo{person}{Chen-Wei Xie}, \bibinfo{person}{Di Chen}, \bibinfo{person}{Feiwu Yu}, \bibinfo{person}{Haiming Zhao}, \bibinfo{person}{Jianxiao Yang}, {et~al\mbox{.}}} \bibinfo{year}{2025}\natexlab{}.
\newblock \showarticletitle{Wan: Open and advanced large-scale video generative models}.
\newblock \bibinfo{journal}{\emph{arXiv preprint arXiv:2503.20314}} (\bibinfo{year}{2025}).
\newblock


\bibitem[Wang and Shi(2023)]%
        {wang2023imagedream}
\bibfield{author}{\bibinfo{person}{Peng Wang} {and} \bibinfo{person}{Yichun Shi}.} \bibinfo{year}{2023}\natexlab{}.
\newblock \showarticletitle{Imagedream: Image-prompt multi-view diffusion for 3d generation}.
\newblock \bibinfo{journal}{\emph{arXiv preprint arXiv:2312.02201}} (\bibinfo{year}{2023}).
\newblock


\bibitem[Wang et~al\mbox{.}(2024)]%
        {wang2024vc}
\bibfield{author}{\bibinfo{person}{Xiang Wang}, \bibinfo{person}{Hangjie Yuan}, \bibinfo{person}{Shiwei Zhang}, \bibinfo{person}{Dayou Chen}, \bibinfo{person}{Jiuniu Wang}, \bibinfo{person}{Yingya Zhang}, \bibinfo{person}{Yujun Shen}, \bibinfo{person}{Deli Zhao}, {and} \bibinfo{person}{Jingren Zhou}.} \bibinfo{year}{2024}\natexlab{}.
\newblock \showarticletitle{{Videocomposer}: Compositional video synthesis with motion controllability}. In \bibinfo{booktitle}{\emph{NeurIPS}}, Vol.~\bibinfo{volume}{36}.
\newblock


\bibitem[Wang et~al\mbox{.}(2019)]%
        {wang2019shape2motion}
\bibfield{author}{\bibinfo{person}{Xiaogang Wang}, \bibinfo{person}{Bin Zhou}, \bibinfo{person}{Yahao Shi}, \bibinfo{person}{Xiaowu Chen}, \bibinfo{person}{Qinping Zhao}, {and} \bibinfo{person}{Kai Xu}.} \bibinfo{year}{2019}\natexlab{}.
\newblock \showarticletitle{Shape2motion: Joint analysis of motion parts and attributes from 3d shapes}. In \bibinfo{booktitle}{\emph{Proceedings of the IEEE/CVF Conference on Computer Vision and Pattern Recognition}}. \bibinfo{pages}{8876--8884}.
\newblock


\bibitem[Wei et~al\mbox{.}(2022)]%
        {wei2022nasam}
\bibfield{author}{\bibinfo{person}{Fangyin Wei}, \bibinfo{person}{Rohan Chabra}, \bibinfo{person}{Lingni Ma}, \bibinfo{person}{Christoph Lassner}, \bibinfo{person}{Michael Zollhoefer}, \bibinfo{person}{Szymon Rusinkiewicz}, \bibinfo{person}{Chris Sweeney}, \bibinfo{person}{Richard Newcombe}, {and} \bibinfo{person}{Mira Slavcheva}.} \bibinfo{year}{2022}\natexlab{}.
\newblock \showarticletitle{Self-supervised Neural Articulated Shape and Appearance Models}. In \bibinfo{booktitle}{\emph{Proceedings IEEE/CVF Conference on Computer Vision and Pattern Recognition (CVPR)}}.
\newblock


\bibitem[Wu et~al\mbox{.}(2022)]%
        {tuneavideo}
\bibfield{author}{\bibinfo{person}{Jay~Zhangjie Wu}, \bibinfo{person}{Yixiao Ge}, \bibinfo{person}{Xintao Wang}, \bibinfo{person}{Stan~Weixian Lei}, \bibinfo{person}{Yuchao Gu}, \bibinfo{person}{Wynne Hsu}, \bibinfo{person}{Ying Shan}, \bibinfo{person}{Xiaohu Qie}, {and} \bibinfo{person}{Mike~Zheng Shou}.} \bibinfo{year}{2022}\natexlab{}.
\newblock \showarticletitle{Tune-A-Video: One-Shot Tuning of Image Diffusion Models for Text-to-Video Generation}.
\newblock \bibinfo{journal}{\emph{arXiv preprint arXiv:2212.11565}} (\bibinfo{year}{2022}).
\newblock


\bibitem[Wu et~al\mbox{.}(2023)]%
        {wu2023lamp}
\bibfield{author}{\bibinfo{person}{Ruiqi Wu}, \bibinfo{person}{Liangyu Chen}, \bibinfo{person}{Tong Yang}, \bibinfo{person}{Chunle Guo}, \bibinfo{person}{Chongyi Li}, {and} \bibinfo{person}{Xiangyu Zhang}.} \bibinfo{year}{2023}\natexlab{}.
\newblock \showarticletitle{Lamp: Learn a motion pattern for few-shot-based video generation}.
\newblock \bibinfo{journal}{\emph{arXiv preprint arXiv:2310.10769}} (\bibinfo{year}{2023}).
\newblock


\bibitem[Wu et~al\mbox{.}(2025)]%
        {wu2025animateanymesh}
\bibfield{author}{\bibinfo{person}{Zijie Wu}, \bibinfo{person}{Chaohui Yu}, \bibinfo{person}{Fan Wang}, {and} \bibinfo{person}{Xiang Bai}.} \bibinfo{year}{2025}\natexlab{}.
\newblock \showarticletitle{AnimateAnyMesh: A Feed-Forward 4D Foundation Model for Text-Driven Universal Mesh Animation}.
\newblock \bibinfo{journal}{\emph{arXiv preprint arXiv:2506.09982}} (\bibinfo{year}{2025}).
\newblock


\bibitem[Xiang et~al\mbox{.}(2020)]%
        {xiang2020sapien}
\bibfield{author}{\bibinfo{person}{Fanbo Xiang}, \bibinfo{person}{Yuzhe Qin}, \bibinfo{person}{Kaichun Mo}, \bibinfo{person}{Yikuan Xia}, \bibinfo{person}{Hao Zhu}, \bibinfo{person}{Fangchen Liu}, \bibinfo{person}{Minghua Liu}, \bibinfo{person}{Hanxiao Jiang}, \bibinfo{person}{Yifu Yuan}, \bibinfo{person}{He Wang}, {et~al\mbox{.}}} \bibinfo{year}{2020}\natexlab{}.
\newblock \showarticletitle{Sapien: A simulated part-based interactive environment}. In \bibinfo{booktitle}{\emph{Proceedings of the IEEE/CVF conference on computer vision and pattern recognition}}. \bibinfo{pages}{11097--11107}.
\newblock


\bibitem[Xie et~al\mbox{.}(2024)]%
        {xie2024sv4d}
\bibfield{author}{\bibinfo{person}{Yiming Xie}, \bibinfo{person}{Chun-Han Yao}, \bibinfo{person}{Vikram Voleti}, \bibinfo{person}{Huaizu Jiang}, {and} \bibinfo{person}{Varun Jampani}.} \bibinfo{year}{2024}\natexlab{}.
\newblock \showarticletitle{Sv4d: Dynamic 3d content generation with multi-frame and multi-view consistency}.
\newblock \bibinfo{journal}{\emph{arXiv preprint arXiv:2407.17470}} (\bibinfo{year}{2024}).
\newblock


\bibitem[Xing et~al\mbox{.}(2024)]%
        {xing2024dynamicrafter}
\bibfield{author}{\bibinfo{person}{Jinbo Xing}, \bibinfo{person}{Menghan Xia}, \bibinfo{person}{Yong Zhang}, \bibinfo{person}{Haoxin Chen}, \bibinfo{person}{Wangbo Yu}, \bibinfo{person}{Hanyuan Liu}, \bibinfo{person}{Gongye Liu}, \bibinfo{person}{Xintao Wang}, \bibinfo{person}{Ying Shan}, {and} \bibinfo{person}{Tien-Tsin Wong}.} \bibinfo{year}{2024}\natexlab{}.
\newblock \showarticletitle{Dynamicrafter: Animating open-domain images with video diffusion priors}. In \bibinfo{booktitle}{\emph{European Conference on Computer Vision}}. Springer, \bibinfo{pages}{399--417}.
\newblock


\bibitem[Yan et~al\mbox{.}(2020)]%
        {yan2020rpm}
\bibfield{author}{\bibinfo{person}{Zihao Yan}, \bibinfo{person}{Ruizhen Hu}, \bibinfo{person}{Xingguang Yan}, \bibinfo{person}{Luanmin Chen}, \bibinfo{person}{Oliver Van~Kaick}, \bibinfo{person}{Hao Zhang}, {and} \bibinfo{person}{Hui Huang}.} \bibinfo{year}{2020}\natexlab{}.
\newblock \showarticletitle{RPM-Net: recurrent prediction of motion and parts from point cloud}.
\newblock \bibinfo{journal}{\emph{arXiv preprint arXiv:2006.14865}} (\bibinfo{year}{2020}).
\newblock


\bibitem[Yang et~al\mbox{.}(2024)]%
        {yang2024cogvideox}
\bibfield{author}{\bibinfo{person}{Zhuoyi Yang}, \bibinfo{person}{Jiayan Teng}, \bibinfo{person}{Wendi Zheng}, \bibinfo{person}{Ming Ding}, \bibinfo{person}{Shiyu Huang}, \bibinfo{person}{Jiazheng Xu}, \bibinfo{person}{Yuanming Yang}, \bibinfo{person}{Wenyi Hong}, \bibinfo{person}{Xiaohan Zhang}, \bibinfo{person}{Guanyu Feng}, {et~al\mbox{.}}} \bibinfo{year}{2024}\natexlab{}.
\newblock \showarticletitle{Cogvideox: Text-to-video diffusion models with an expert transformer}.
\newblock \bibinfo{journal}{\emph{arXiv preprint arXiv:2408.06072}} (\bibinfo{year}{2024}).
\newblock


\bibitem[Yao et~al\mbox{.}(2025)]%
        {yao2025sv4d}
\bibfield{author}{\bibinfo{person}{Chun-Han Yao}, \bibinfo{person}{Yiming Xie}, \bibinfo{person}{Vikram Voleti}, \bibinfo{person}{Huaizu Jiang}, {and} \bibinfo{person}{Varun Jampani}.} \bibinfo{year}{2025}\natexlab{}.
\newblock \showarticletitle{Sv4d 2.0: Enhancing spatio-temporal consistency in multi-view video diffusion for high-quality 4d generation}.
\newblock \bibinfo{journal}{\emph{arXiv preprint arXiv:2503.16396}} (\bibinfo{year}{2025}).
\newblock


\bibitem[Yu et~al\mbox{.}(2023)]%
        {yu2023magvit}
\bibfield{author}{\bibinfo{person}{Lijun Yu}, \bibinfo{person}{Yong Cheng}, \bibinfo{person}{Kihyuk Sohn}, \bibinfo{person}{Jos{\'e} Lezama}, \bibinfo{person}{Han Zhang}, \bibinfo{person}{Huiwen Chang}, \bibinfo{person}{Alexander~G Hauptmann}, \bibinfo{person}{Ming-Hsuan Yang}, \bibinfo{person}{Yuan Hao}, \bibinfo{person}{Irfan Essa}, {et~al\mbox{.}}} \bibinfo{year}{2023}\natexlab{}.
\newblock \showarticletitle{Magvit: Masked generative video transformer}. In \bibinfo{booktitle}{\emph{Proceedings of the IEEE/CVF Conference on Computer Vision and Pattern Recognition}}. \bibinfo{pages}{10459--10469}.
\newblock


\bibitem[Zeng et~al\mbox{.}(2024)]%
        {zeng2024stag4d}
\bibfield{author}{\bibinfo{person}{Yifei Zeng}, \bibinfo{person}{Yanqin Jiang}, \bibinfo{person}{Siyu Zhu}, \bibinfo{person}{Yuanxun Lu}, \bibinfo{person}{Youtian Lin}, \bibinfo{person}{Hao Zhu}, \bibinfo{person}{Weiming Hu}, \bibinfo{person}{Xun Cao}, {and} \bibinfo{person}{Yao Yao}.} \bibinfo{year}{2024}\natexlab{}.
\newblock \showarticletitle{Stag4d: Spatial-temporal anchored generative 4d gaussians}. In \bibinfo{booktitle}{\emph{European Conference on Computer Vision}}. Springer, \bibinfo{pages}{163--179}.
\newblock


\bibitem[Zhang et~al\mbox{.}(2024)]%
        {zhang20244diffusion}
\bibfield{author}{\bibinfo{person}{Haiyu Zhang}, \bibinfo{person}{Xinyuan Chen}, \bibinfo{person}{Yaohui Wang}, \bibinfo{person}{Xihui Liu}, \bibinfo{person}{Yunhong Wang}, {and} \bibinfo{person}{Yu Qiao}.} \bibinfo{year}{2024}\natexlab{}.
\newblock \showarticletitle{4diffusion: Multi-view video diffusion model for 4d generation}.
\newblock \bibinfo{journal}{\emph{Advances in Neural Information Processing Systems}}  \bibinfo{volume}{37} (\bibinfo{year}{2024}), \bibinfo{pages}{15272--15295}.
\newblock


\bibitem[Zhang et~al\mbox{.}(2018)]%
        {zhang2018perceptual}
\bibfield{author}{\bibinfo{person}{Richard Zhang}, \bibinfo{person}{Phillip Isola}, \bibinfo{person}{Alexei~A Efros}, \bibinfo{person}{Eli Shechtman}, {and} \bibinfo{person}{Oliver Wang}.} \bibinfo{year}{2018}\natexlab{}.
\newblock \showarticletitle{The Unreasonable Effectiveness of Deep Features as a Perceptual Metric}. In \bibinfo{booktitle}{\emph{CVPR}}.
\newblock


\bibitem[Zhao et~al\mbox{.}(2024)]%
        {zhao2024motiondirector}
\bibfield{author}{\bibinfo{person}{Rui Zhao}, \bibinfo{person}{Yuchao Gu}, \bibinfo{person}{Jay~Zhangjie Wu}, \bibinfo{person}{David~Junhao Zhang}, \bibinfo{person}{Jia-Wei Liu}, \bibinfo{person}{Weijia Wu}, \bibinfo{person}{Jussi Keppo}, {and} \bibinfo{person}{Mike~Zheng Shou}.} \bibinfo{year}{2024}\natexlab{}.
\newblock \showarticletitle{Motiondirector: Motion customization of text-to-video diffusion models}. In \bibinfo{booktitle}{\emph{European Conference on Computer Vision}}. Springer, \bibinfo{pages}{273--290}.
\newblock


\bibitem[Zhou et~al\mbox{.}(2022)]%
        {zhou2022magicvideo}
\bibfield{author}{\bibinfo{person}{Daquan Zhou}, \bibinfo{person}{Weimin Wang}, \bibinfo{person}{Hanshu Yan}, \bibinfo{person}{Weiwei Lv}, \bibinfo{person}{Yizhe Zhu}, {and} \bibinfo{person}{Jiashi Feng}.} \bibinfo{year}{2022}\natexlab{}.
\newblock \showarticletitle{Magicvideo: Efficient video generation with latent diffusion models}.
\newblock \bibinfo{journal}{\emph{arXiv preprint arXiv:2211.11018}} (\bibinfo{year}{2022}).
\newblock


\end{thebibliography}

\clearpage

\section{Algorithm} Here, we explain in detail the algorithm used to retrieve the motion axis and origin from the point cloud. In this paper, we consider two types of motion, 1) Revolute Motion 2) Prismatic Motion. 
\\
\emph{Revolute Motion.} Given a set of vertices of a mesh  $\mathcal{M}$ the revolute motion is represented as: 
\begin{equation}
    \Vec{x}^{*} = \mathbf{R} \ (\Vec{x} - \Vec{o}) + \Vec{o}
    \label{eq: revolute_motion}
\end{equation}
where, $\Vec{x}$ is the original mesh vertex, $\Vec{x}^{*}$ is the transformed mesh vertex after articulation, $\Vec{o}$ is motion axis origin, $\mathbf{R}$ is the rotation matrix around the predicted motion axis direction, $\Vec{a}=[a_{x},\ a_{y},\ a_{z}]$. Here, $\|\Vec{a}\|=1$. Given an angle $\theta$, with which we want to rotate around the motion axis, we can compute the rotation matrix using Rodrigues formula as,
\begin{gather}
\mathbf{K} =
\begin{bmatrix}
0 & -\text{a}_z & \text{a}_y \\
\text{a}_z & 0 & -\text{a}_x \\
-\text{a}_y & \text{a}_x & 0
\end{bmatrix}, \\
\mathbf{R} = \mathbf{I} + \sin(\theta) \ \mathbf{K} + (1 - \cos(\theta)) \ \mathbf{K}^2
\end{gather}
\emph{Prismatic Motion.} Unlike revolute motion, prismatic motion is defined using only motion axis, $\Vec{a}$, where $\|\Vec{a}\|=1$. Given a motion range $[M_{max}, M_{min}]$ within which we would like to articulate a part, the magnitude $\gamma$ and ultimately the transformed vertex of the mesh $\mathcal{M}$ is defined as:
\begin{equation}
    \Vec{a}^{*} = \gamma * \Vec{a}, \quad \Vec{x}^{*} = \Vec{x} + \Vec{a}^{*}
    \label{eq: prismatic_motion}
\end{equation}
where, $\gamma$ is the magnitude of scaling computed from the motion range, $\Vec{x}$ is the original mesh vertex and $\Vec{x}^{*}$ is the transformed mesh vertex. The algorithm to find motion parameters is described in Algorithm \ref{alg:find_motion_parameters}.

\begin{algorithm}[hbt!]
    \SetCommentSty{commfont}
    \SetKwInOut{Input}{input}\SetKwInOut{Output}{output}
    \Input{Static Mesh $\mathcal{M}$, Point Clouds $P_{N_{f}}$, Motion Type $M_{type}$, Segmentation Labels $L$, Instance ID $y$}
    \Output{Motion Axis $\Vec{a}$, Motion Origin $\Vec{o}$} 
    \BlankLine
    {$\Vec{a}, \Vec{o} \leftarrow \Vec{0}, \Vec{0}$}\;
    \textcolor{blue}{\tcp{Find $OBB$, where $OBB = (\Vec{v_{1}},\Vec{v_{2}},\Vec{v_{3}},\Vec{c},d_{v_{1}},d_{v_{2}},d_{v_{3}})$}}
    $OOB \leftarrow$ \texttt{Find\_Oriented\_Bounding\_Box}$(\mathcal{M}, \ y)$ \;
    $A_{hyp} \leftarrow (\Vec{v_{1}},\Vec{v_{2}},\Vec{v_{3}},-\Vec{v_{1}},-\Vec{v_{2}},-\Vec{v_{3}})$\; 
    $O_{hyp} \leftarrow (\Vec{c} \pm 0.5 \ (d_{v_{1}}.\Vec{v_{1}}), \Vec{c} \pm 0.5(d_{v_{2}}.\Vec{v_{2}}),\Vec{c} \pm 0.5(d_{v_{3}}.\Vec{v_{3}}), \Vec{c})$\;
    $min\_dist \leftarrow 0.0$\;  
    \For{each $\Vec{a_{i}}$ in $A_{hyp}$ }{
    \For{each $\Vec{o_{i}}$ in $O_{hyp}$}{
        \textcolor{blue}{\tcp{transform vertices of $\mathcal{M}$ as per $M_{type}$ \& find best match.}}
        \If{$M_{type}$ = "revolute"}
        {
            
            Mesh $\mathcal{M}^{'} \leftarrow$ \texttt{Revolute}($\mathcal{M}$, $\Vec{a_{i}}$, $\Vec{o_{i}}$, y) \tcp{Eq: \ref{eq: revolute_motion}}
        }
        \If{$M_{type}$ = "prismatic"}
        {
            Mesh $\mathcal{M}^{'} \leftarrow$ \texttt{Prismatic}($\mathcal{M}$, $\Vec{a_{i}}$, y) \tcp{Eq: \ref{eq: prismatic_motion}}
        }
        $P_{\mathcal{M}} \leftarrow$ $\mathcal{M}.$vertices\;
        $e \leftarrow \texttt{chamfer\_distance($P_{\mathcal{M}}$,$P_{N_{f}}$)}$\;
        \If{\texttt{e} < $min\_dist$}
        {
            $\Vec{a} \leftarrow \Vec{a_{i}}$;  $\Vec{o} \leftarrow \Vec{o_{i}}$
        }
            
        }
        
    }
    \caption{Find\_Motion\_Parameters()}
    \label{alg:find_motion_parameters}
\end{algorithm}

\begin{figure*}
    \includegraphics[width=\textwidth]{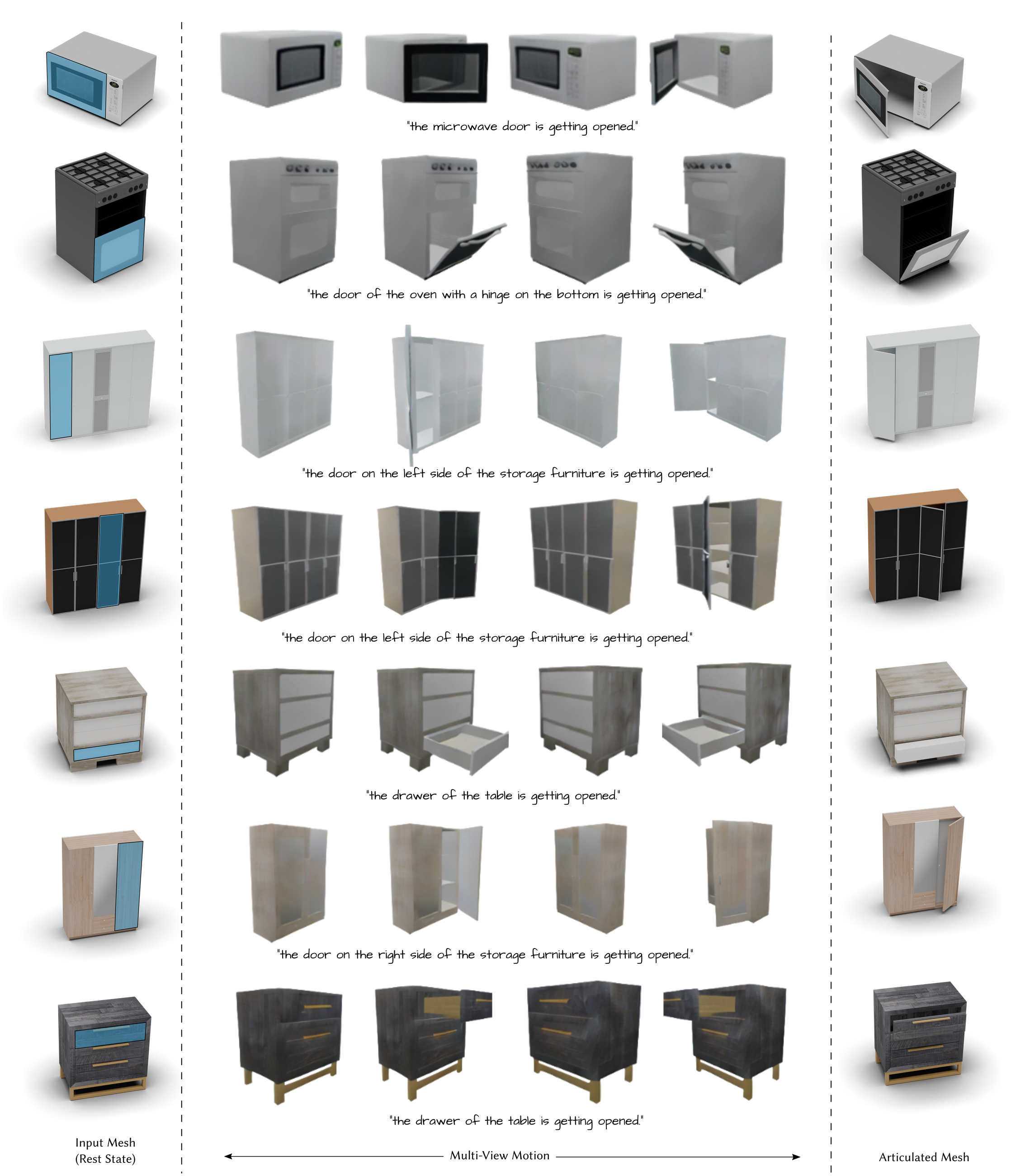}

\caption{Qualitative results of generalization of Multi-view video generation and 3D Motion Axis prediction on ACD dataset \cite{iliash2024s2o}.}
\label{fig:acd_new_1}
\end{figure*}

\begin{figure*}
    \includegraphics[width=\textwidth]{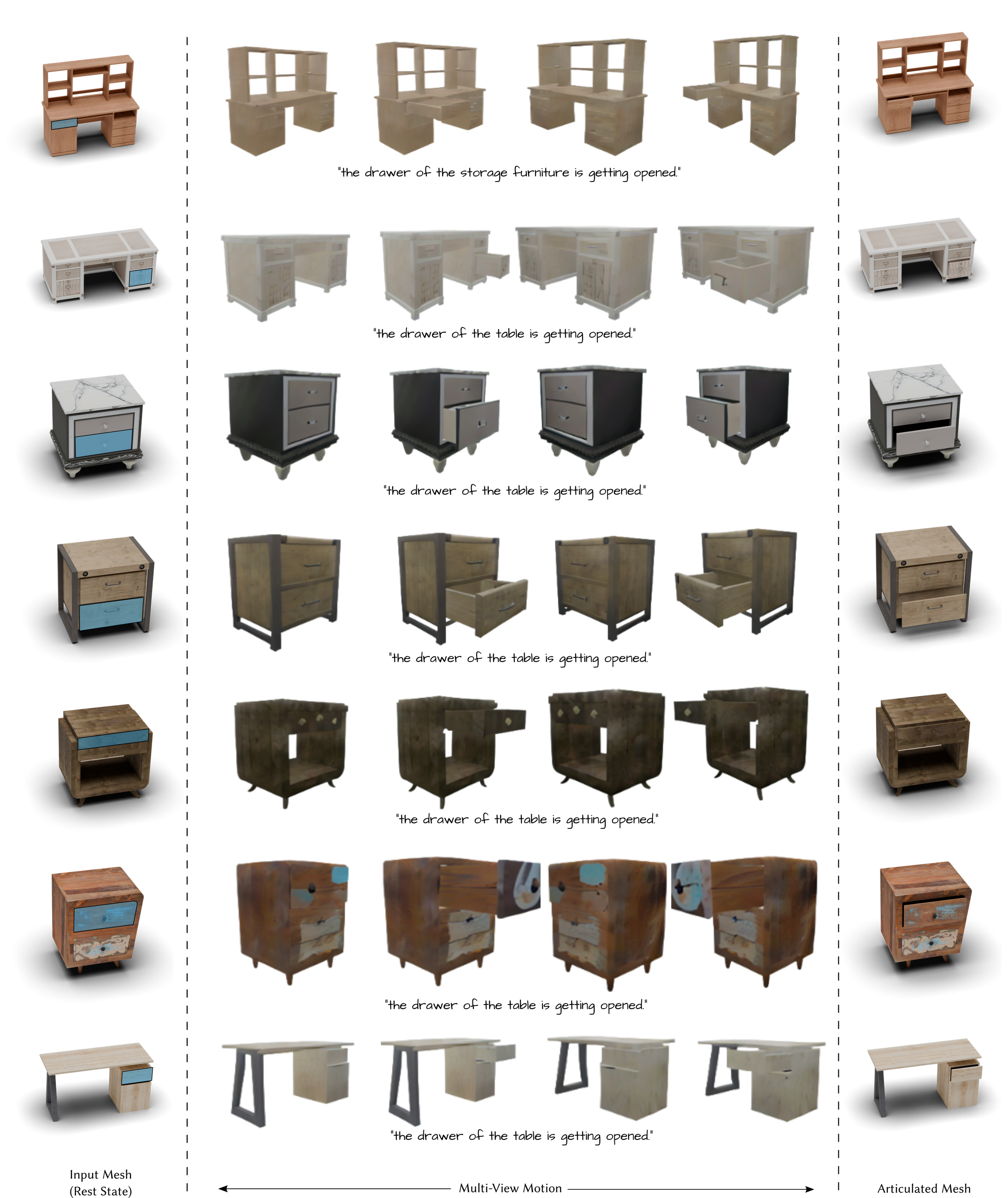}

\caption{Qualitative results of generalization of Multi-view video generation and 3D Motion Axis prediction on ACD dataset \cite{iliash2024s2o}.}
\label{fig:acd_new_2}
\end{figure*}

\begin{figure*}
    \includegraphics[width=\textwidth]{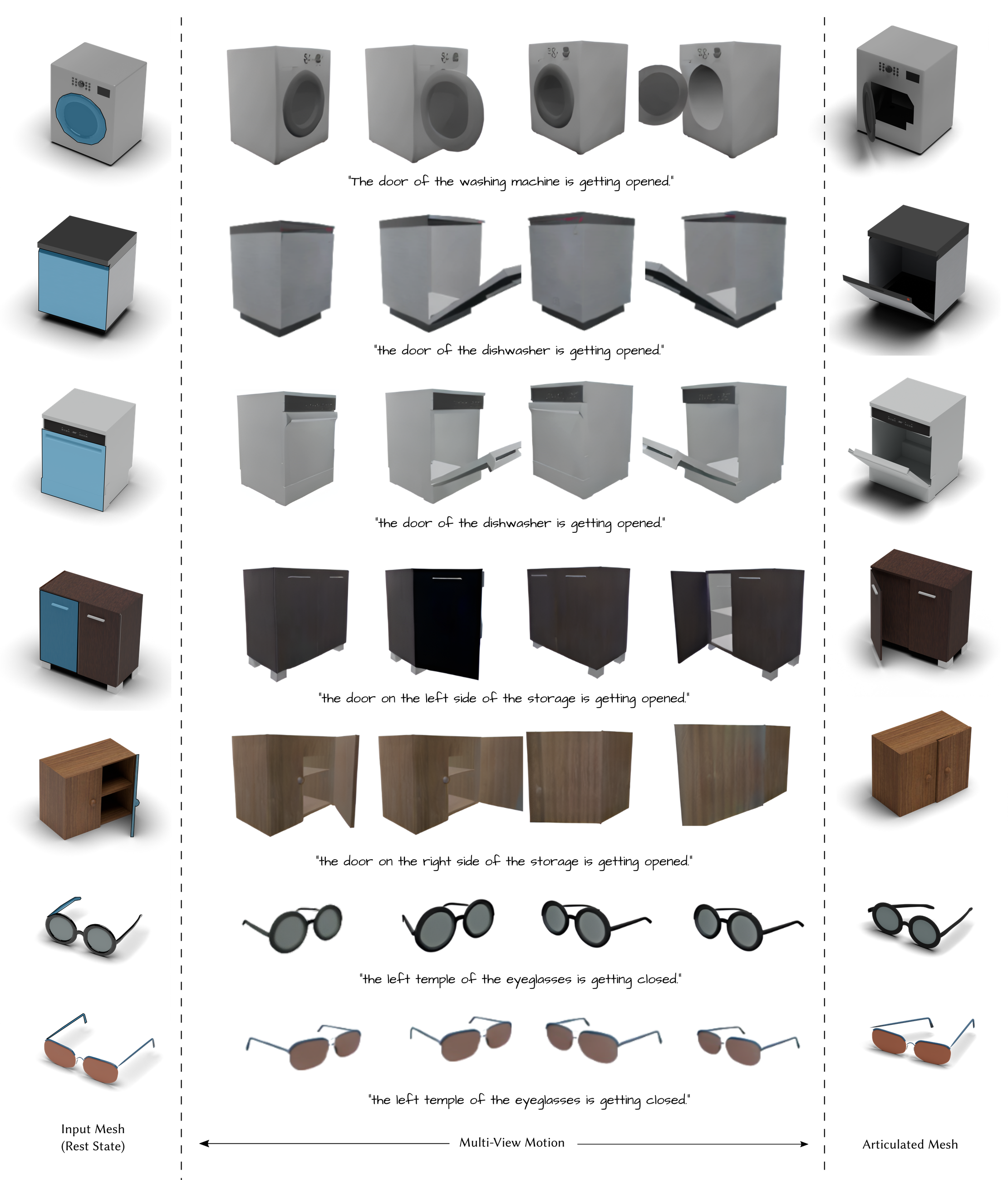}

\caption{Qualitative results of generalization of Multi-view video generation and 3D Motion Axis prediction on Objaverse dataset \cite{deitke2023objaverse, deitke2024objaverse}.}
\label{fig:objaverse_new_3}
\end{figure*}

\begin{figure*}
\centering
    \includegraphics[scale=0.23]{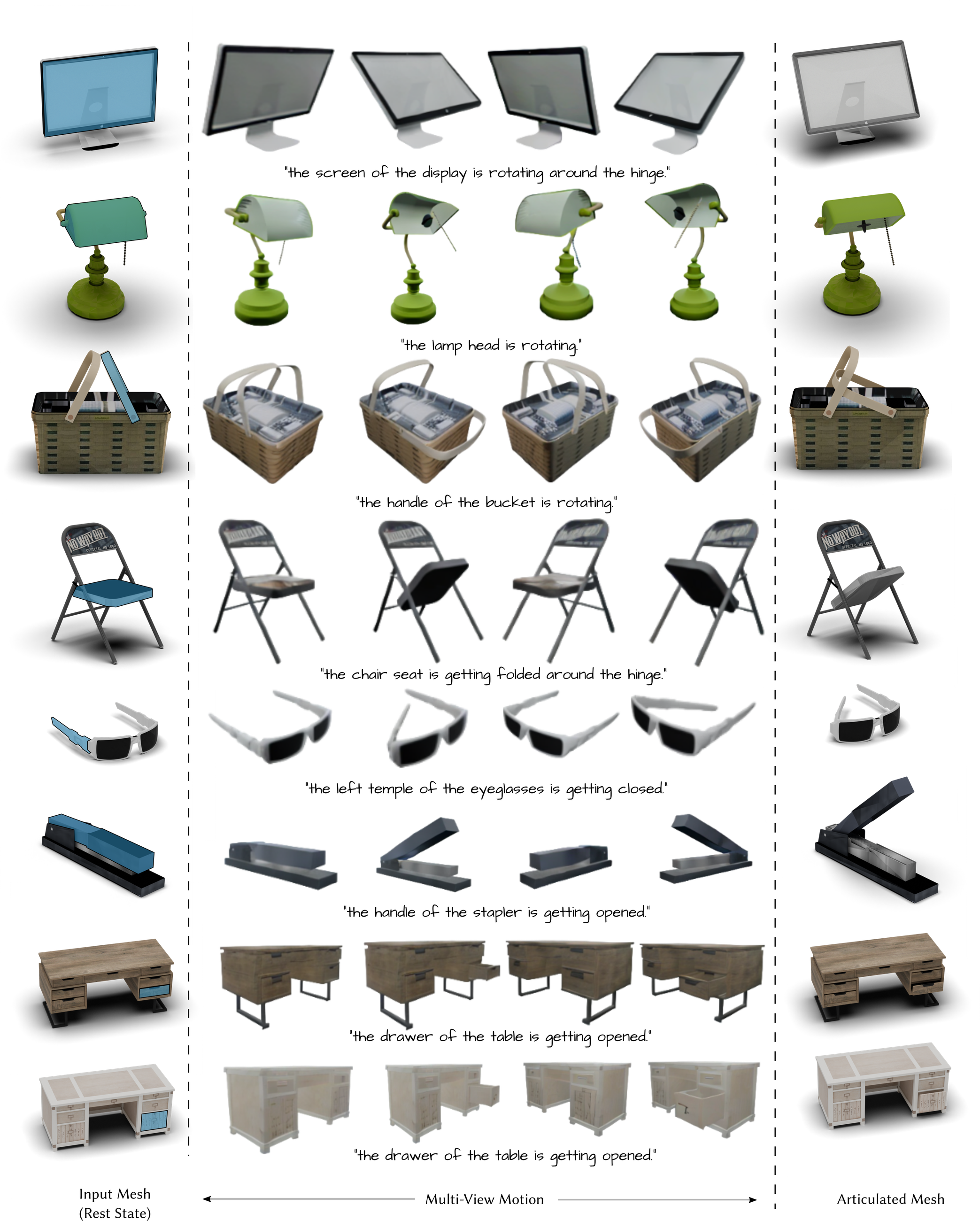}

\caption{Qualitative results of generalization of Multi-view video generation and 3D Motion Axis prediction on PartNet-Mobility and ACD datasets.}
\label{fig:ps_new_4}
\end{figure*}

\end{document}